\documentclass[10pt,twocolumn,letterpaper]{article}

\usepackage{cvpr}
\usepackage{times}
\usepackage{epsfig}
\usepackage{graphicx}
\usepackage{amsmath}
\usepackage{amssymb}
\usepackage{color}
\usepackage{multirow}
\usepackage{diagbox}
\usepackage{booktabs}
\usepackage{makecell}
\usepackage[linesnumbered,ruled]{algorithm2e}
\usepackage{algpseudocode}
\usepackage{amsmath}
\usepackage{bm}

\usepackage[pagebackref=true,breaklinks=true,letterpaper=true,colorlinks,bookmarks=false]{hyperref}

 \cvprfinalcopy % *** Uncomment this line for the final submission

\def\cvprPaperID{****} % *** Enter the CVPR Paper ID here

\ifcvprfinal\fi
\begin{document}

%%%%%%%%% TITLE
\title{A Structural Correlation Filter Combined with A Multi-task Gaussian Particle Filter for Visual Tracking}

\author{Manna Dai\\
{\tt\small Manna.Dai@student.uts.edu.au}
% For a paper whose authors are all at the same institution,
% omit the following lines up until the closing ``}''.
% Additional authors and addresses can be added with ``\and'',
% just like the second author.
% To save space, use either the email address or home page, not both
\and
Shuying Cheng\\
{\tt\small sycheng@fzu.edu.cn}
\and
Xiangjian He\\
{\tt\small Xiangjian.He@uts.edu.au}
\and
Dadong Wang\\
{\tt\small Dadong.Wang@data61.csiro.au}
}

\maketitle
%\thispagestyle{empty}

%%%%%%%%% ABSTRACT
\begin{abstract}
   In this paper, we propose a novel structural correlation filter combined with a multi-task Gaussian particle filter (KCF-GPF) model for robust visual tracking. We first present an assemble structure where several KCF trackers as weak experts provide a preliminary decision for a Gaussian particle filter to make a final decision. The proposed method is designed to exploit and complement the strength of a KCF and a Gaussian particle filter. Compared with the existing tracking methods based on correlation filters or particle filters, the proposed tracker has several advantages. First, it can detect the tracked target in a large-scale search scope via weak KCF trackers and evaluate the reliability of weak trackers\rq decisions for a Gaussian particle filter to make a strong decision, and hence it can tackle fast motions, appearance variations, occlusions and re-detections. Second, it can effectively handle large-scale variations via a Gaussian particle filter. Third, it can be amenable to fully parallel implementation using importance sampling without resampling, thereby it is convenient for VLSI implementation and can lower the computational costs. Extensive experiments on the OTB-2013 dataset containing 50 challenging sequences demonstrate that the proposed algorithm performs favourably against 16 state-of-the-art trackers.
\end{abstract}

%%%%%%%%% BODY TEXT
\section{Introduction}

Visual tracking is one of the most fundamental problems in computer vision due to its numerous applications such as video surveillance, motion analysis, vehicle navigation and human computer interactions. Although a great progress has been seen on developing algorithms \cite{Henriques2015High,Valmadre2017End,Wang2017Large,Zhang2016Robust} and benchmark evaluations \cite{Wu2013Online} for visual tracking, visual tracking is still a challenging problem in the situations of heavy illumination changes, pose deformations, partial and full occlusions, large scale variations, background clutter and fast motion.

Correlation filters have recently attracted a great attention due to their rapid speeds of calculation and robust tracking performance \cite{Li2015Reliable,Liu2016Structural,Liu2015Real,Ma2015Hierarchical}.
Bolme et al. \cite{Bolme2010Visual} proposed an adaptive correlation filer, called MOSSE, for producing ASEF-like filters by fewer training images. Henriques et al. \cite{Rui2012Exploiting} extended the correlation-filter-based trackers to kernel-based training, called CSK method, to utilize a circulant structure of one image patch to conduct dense sampling, and then improved the KCF tracker \cite{Henriques2015High} by using multi-channel inputs and HOG descriptors. Danelljan et al. \cite{Danelljan2016Discriminative} developed the DSST method handling scale changes of a target, and Choi et al. \cite{Choi2016Visual} proposed a spatially attentional weight map to weight various correlation filters. Ma et al. \cite{Ma2015Long} used a correlation filter as a short-term tracker and an online random fern classifier for re-detection as a long-term memory system.

Although achieved the appealing results both in precision and success rate, these correlation-filter-based trackers cannot deal with fast motions and scale variations well. For example, although two correlation-filter-based trackers, namely SCT\cite{Choi2016Visual} and KCF \cite{Henriques2015High}, have achieved state-of-the-art results and have beaten all other attended trackers in terms of accuracy in the OTB-2013 dataset \cite{Wu2013Online}, they fail to track a target object when partial occlusions or illumination changes occur.

To deal with the above issues, we propose a novel and an ensemble tracker, which first builds weak trackers by applying structural correlation filters, and then integrate all weak trackers into one stronger tracker using a reliability evaluation and a multi-task Gaussian particle filter. Each weak tracker is treated as an expert and the weights of all experts are computed via their confidence maps. Through the reliability evaluation, the weak trackers provide weak decisions for the Gaussian particle filter to make a final decision. The tracking result in the current frame is interred by the weights of the Gaussian particle filter. Particles respectively calculate two results using HOG and gray-normalization features, and therefore the Gaussian particle filter conducts a multi-task tracking for making a strong decision.

The contributions in this paper are summarized as follows.
\begin{itemize}
  \item A novel ensemble algorithm is proposed to combine structural correlation filters and a Gaussian particle filter into a single stronger tracker.
  \item A large-scale search scope algorithm using for multiple structural correlation filters is proposed,  considering spatial geometric relations between target locations in consecutive frames and therefore making the search reliable.
  \item Extensive experiments in the OTB-2013 benchmark dataset \cite{Wu2013Online} with 50 challenging sequences and 11 various attributes to demonstrate the outperformance of the proposed method in comparison with 16 state-of-the-art trackers.
\end{itemize}

%-------------------------------------------------------------------------
\section{Related Work}

A comprehensive tracking review can be found in \cite{Wu2013Online,Yilmaz2006Object,Smeulders2014Visual}. In this section, we discuss the methods closely related to this work, mainly regarding multi-task correlation filters, Gaussian particle filters and Average Peak-to-Correlation Energy (APCE).

\subsection{Structural Correlation Filters}

 Qi et al. \cite{Qi2016Hedged} described the weak correlation filers on CNN features in each layer and Liu et al. \cite{Liu2016Structural} proposed the concept of the structural correlation filter.

In \cite{Qi2016Hedged}, $\mathbf{\emph{X}}^{k}\in \mathbb{R}^{P\times Q\times D}$ denote the feature map extracted from the $k$-th convolutional layer with Gaussian function label $\mathbf{\emph{Y}}\in \mathbb{R}^{P\times Q}$. Let $\mathbf{\mathcal{X}}^{k}=\mathcal{F}(\mathbf{\emph{X}}^{k})$ and $\mathbf{\mathcal{Y}}=\mathcal{F}(\mathbf{\emph{Y}})$, where $\mathcal{F}(\cdot)$ represents the discrete Fourier transformation (DFT).
The objective function of correlation filter method \cite{Qi2016Hedged} can be extended into its $k$-th filter modeled as
\begin{equation}\label{Eq:1}
\mathbf{\mathcal{W}}^{k}=\mathrm{arg} \min_{\mathbf{\mathcal{W}}}\|\mathbf{\mathcal{Y}}-\mathbf{\mathcal{X}}^{k}\cdot \mathbf{\mathcal{W}}\|^{2}_{F} + \lambda \| \mathbf{\mathcal{W}}\|^{2}_{F},
\end{equation}
where
\begin{equation}\label{Eq:2}
  \mathbf{\mathcal{X}}^{k}\cdot \mathbf{\mathcal{W}} = \sum_{d=1}^{D}\mathbf{\mathcal{X}}^{k}_{*,*,d}\bigodot\mathbf{\mathcal{W}}_{*,*,d}.
\end{equation}
Here, the symbol $\bigodot$ is the element-wise product.

The optimization problem in Eq. \ref{Eq:1} has a simple closed form solution, which can be efficiently computed in the Fourier domain by
\begin{equation}\label{Eq:3}
  \mathbf{\mathcal{W}}^{k}_{*,*,d} = \frac{\mathbf{\mathcal{Y}}}{\mathbf{\mathcal{X}}^{k}\cdot \mathbf{\mathcal{X}}^{k} + \lambda}\bigodot \mathbf{\mathcal{X}}^{k}_{*,*,d}.
\end{equation}

Given the testing data $\mathbf{\emph{T}}^{k}$ from the output of the $k$-th layer, we first transform it to the Fourier domain $\mathbf{\mathcal{T}}^{k}=\mathcal{F}(\mathbf{\emph{T}}^{k})$, and then the responses can be computed by
\begin{equation}\label{Eq:4}
\mathbf{\emph{S}}^{k}=  \mathcal{F}^{-1}(\mathbf{\mathcal{T}}^{k}\cdot\mathbf{\mathcal{W}}^{k}),
\end{equation}
where $\mathcal{F}^{-1}$ denotes the inverse of DFT.

The $k$-th weak tracker outputs the target position with the largest response
\begin{equation}\label{Eq:5}
(x^{k},y^{k})=\mathrm{arg}\max_{x',y'} \mathbf{\emph{S}}^{k}(x',y').
\end{equation}

\subsection{Gaussian Particle Filters (GPF)}

Kotecha and Djuric \cite{Kotecha2003Gaussian} introduced the Gaussian Particle Filter (GPF), which is used for tracking filtering and predictive distributions encountered in Dynamic State-Space models (DSS) \cite{Harrison1976Bayesian}. The DSS model represents the time-varying dynamics of an unobserved state variable. GPF is based on the particle filtering and Gaussian filtering concepts. Gaussian filters provide Gaussian approximations to the filtering and predictive distributions, and they include Extended Kalman Filter (EKF) \cite{Julier2004Unscented} and its
variations \cite{Jazwinski1970Stochastic,Mendel1980Optimal,Sorenson1988Recursive,Welch2010An}. Unlike EKF, which assumes that predictive distributions are Gaussian and employs linearization of the functions in the process and observation equations, GPF updates the Gaussian approximations using particles. GPF only propagates the posterior mean and covariance of an unobserved state variable in a DSS model, and essentially importance sampling makes the procedure simple.

Particle filters \cite{Gustafsson2002Particle,Merwe2001The,Nummiaro2003An} use sequential importance sampling (SIS) \cite{Liu2001A} to update the posterior distributions. GPF is quite similar to SIS filters by the fact that importance sampling is used to obtain particles. However, a phenomenon called sample degeneration occurs wherein only a few particles representing the distribution have significant weights. A procedure called resampling \cite{Liu1998Sequential} has been introduced to mitigate this problem, but it may give limited results and may be computationally expensive. Since GPF approximates posterior distributions as Gaussians, unlike the SIS filters, particle resampling is not required. This results in a reduced complexity of GPF as compared with SIS with resampling and is a major advantage. Furthermore, Berzuini et al. \cite{Carlo1997Dynamic} reported that particle filters with resampling also had bias due to resampling, and resampling in SIS filters is a nonparallel operation. Fortunately, resampling would never occur in GPF simulation examples, and the particle filters in GPF are amenable to parallel implementation. Therefore, GPF is more amenable for fully parallel implementation in very large scale integration (VLSI) than SIS.

Simulation results are presented to demonstrate the versatility and improved performance of GPF over conventional Gaussian filters and the lower complexity of GPF than the known particle filters. However, the parallelizibility of GPF and the absence of resampling makes it convenient for VLSI implementation and, hence, feasible for practical real-time applications.

\subsection{Ensemble trackers}

Multiple component trackers have been combined with hand-crafted features to develop ensemble tracking methods \cite{Avidan2007Ensemble,Bai2013Randomized,Wang2014Ensemble} for visual tracking. For example, several ensemble methods \cite{Avidan2007Ensemble,Bai2013Randomized} using a boosting framework \cite{Freund1995A} train each component weak tracker to classify foreground objects and background. In \cite{Wang2014Ensemble}, Wang and Yeung used a conditional particle filter to infer a target's position and the reliability of each component tracker. Qi et al. \cite{Qi2016Hedged} treated tracking as a decision-theoretic online learning task and the tracked target was inferred by using the decisions from multiple expert trackers. Similar to \cite{Qi2016Hedged}, we consider visual tracking as a decision-theoretic online learning task \cite{Chaudhuri2009A}, and use it in the structure of multiple correlation filters combined with a Gaussian particle filter. That is, in every round, each correlation filter makes a decision and the final decision is determined by a Gaussian particle filter.

\subsection{Average Peak-to-Correlation Energy (APCE)}

For correlation filters, the peak value $F_{max}$ denotes the maximum response score of a response map. To measure the fluctuation degree of a response map and the reliability degree of a detected object, Wang et al. \cite{Wang2017Large} proposed an average peak-to-correlation energy (APCE) which is defined as
\begin{equation}\label{Eq:6}
  APCE = \frac{|F_{max}-F_{min}|^{2}}{\mathrm{mean}\left( \sum\limits_{w,h} (F_{w,h}-F_{min})^{2}  \right)}
\end{equation}
where $F_{max}$, $F_{min}$ and $F_{w,h}$ denote the maximum, minimum
and the $w$-th row and $h$-th column elements of $F(s,y;w)$.
APCE indicates the fluctuated degrees of response maps and the confidence degrees of the tracked results. For sharper peaks and less noise, in the case that the target fully appearing in a tracking region, APCE will become greater and the response map will become smoother except for only one sharp peak. On the other hand, APCE will be small if an object is occluded or missing.

\begin{figure*}
  \centering
  % Requires \usepackage{graphicx}
  \includegraphics[width=5.5in]{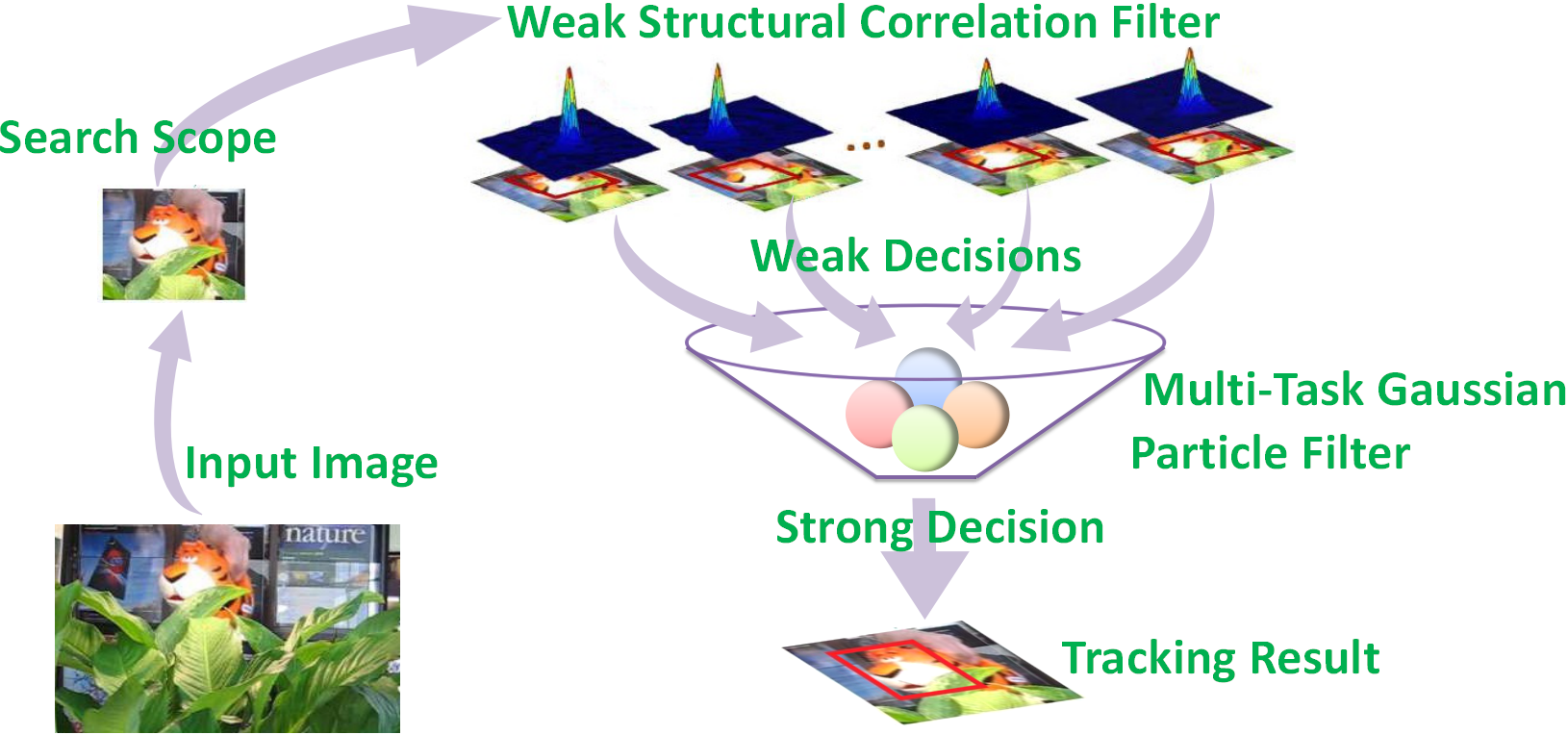}\\
  \caption{Flowchart of the proposed algorithm. The proposed algorithm consists of three components: 1) constructing multiple weak trackers using correlation filters where each one is trained using HOG features and makes a weak decision (Section \ref{Weak Structural Correlation Filter}); 2) evaluating the reliability degree of each weak decision via maximum response score and APCE measure and the most reliable decision is used for the next step (Section \ref{Reliability Degree of Correlation Filter}); 3) constructing a stronger tracker and making a final decision via a multi-task Gaussian particle filter which takes the most reliable decision into account (Section \ref{Multi-task Gaussian Particle Filter}).}\label{fig:flowchart}
\end{figure*}

%-------------------------------------------------------------------------
\section{Proposed Algorithm}

In this section, we present the combination of structural correlation filters with a multi-task Gaussian particle filter for ensemble tracking, namely KCF-GPF. Different from the KCF method \cite{Henriques2015High,Rui2012Exploiting} that learns a single correlation filter in a fixed-size area, KCF-GPF is proposed to construct multiple weak correlation filters in a more reliable search scope for dealing with fast motion issues and bound effects in the conventional correlation filters. The Gaussian particle filter jointly learns particle weights based on different features to make a stronger tracker by a multi-task method. Furthermore, our tracker can effectively handle scale variations via the sampling strategy of a Gaussian particle filter. Overall, the proposed ensemble method will achieves the following two goals: 1) weak expert trackers are tuned to separate a foreground object from background and 2) the ensemble as a whole ensures the temporal coherence of each part of the tracker.

\subsection{Weak Structural Correlation Filter}
\label{Weak Structural Correlation Filter}

A conventional correlation filter has bound effects \cite{Danelljan2015Learning} during a target tracking and it interferes with the progress of target detection with fast motions. For this, we extend the conventional search window for a single tracker to a large-scale search scope for multiple trackers, and exploit spatial-geometric relations between target locations in consecutive frames to make the search scope reliable.

\begin{figure}
  \centering
  \includegraphics[width=3in]{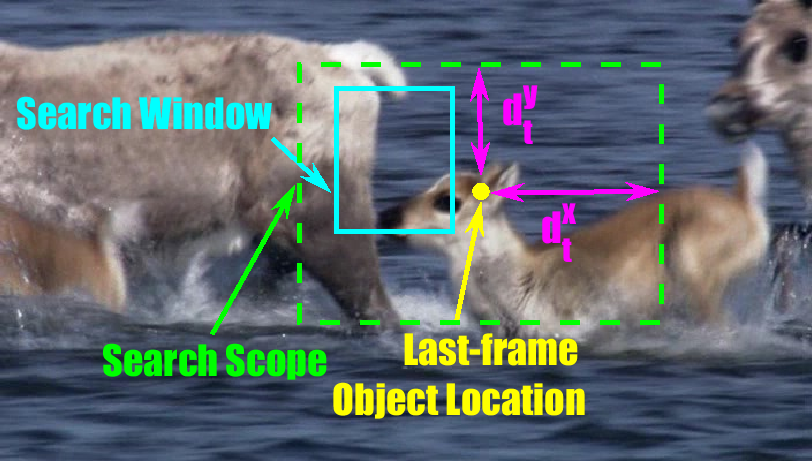}\\
  \caption{Illustration of the target tracking using the weak structural correlation filter in sequence \textbf{Deer} from  OTB-2013 dataset \cite{Wu2013Online}. The search scope is based on the max historical moving distance of the object in horizontal and vertical directions at time $t$, namely $d_{t}^{x}$ and $d_{t}^{y}$, and considers the last-frame object location as the center. A search window slides in the search scope to enclose a sample corresponding to a weak correlation filter.}\label{Fig:1}
\end{figure}

The feature map $\mathbf{\emph{X}}^{k}$ in Eq. \ref{Eq:1} is extracted from an image patch which is sampled by a search sliding window (see Figure. \ref{Fig:1}), where $d_{t}$ donates the maximum historical-moving-distance of a tracked target in Eq. \ref{Eq:17} at time $t$ and the object location in the previous frame is the center of the search scope. Besides it, we respectively use $d_{t}^{x}$ and $d_{t}^{y}$ to represent the horizontal and vertical moving distances at time $t$.

Given the initial confidence weights of all weak experts, in the current round, a further decision is made based on the expert tracker with the greatest weight. In the visual tracking scenario, it is natural to treat each KCF tracker as an expert and then predict the target position at time $t$ by
\begin{equation}\label{Eq:7}
  (x^{*}_{t},y^{*}_{t})= (x^{k}_{t},y^{k}_{t}) \cdot [1-\mathrm{sign}(\left|w^{k}_{t}-\max w^{k}_{t}\right|)],
\end{equation}
where $w^{k}_{t}$ is the weight of expert $k$ and $\sum_{k=1}^{K}w^{k}_{t}=1$, $\left|\cdot \right|$ denotes the absolute value and $\mathrm{sign}$ represents the signum function.

\textbf{Kernel Selection:}  We choose the Gaussian kernel in the existing correlation filter tracker \cite{Henriques2015High}.

\textbf{Feature Representation:} Similar to \cite{Henriques2015High}, we use HOG features with 31 bins. However, our tracker is quite generic and any dense feature representation with arbitrary dimensions can be incorporated.

Compared to the HDT \cite{Qi2016Hedged} and SCF \cite{Liu2016Structural} methods which are similar to the proposed weak structural correlation filter, we demonstrate differences among these approaches as follows.
\begin{enumerate}
  \item HDT uses CNN features, while SCF and KCF-GPF are based on HOG features.
  \item The features of HDT are extracted from one layer to build a weak tracker, and the part-based correlation filter SCF samples several parts of a target object to construct features, while KCF-GPF samples via sliding window in a search scope which is subject to the maximum historic-moving-displacement of the tracked target over time $t$.
  \item  In HDT, the target position is made based on the weighted decisions of all experts and SCF solves the optimization problem using the fast first-order Alternating Direction Method of Multipliers (ADMM) \cite{Boyd2011Distributed}, while KCF-GPF exploits Eq. \ref{Eq:7} to infer the ultimate target position.
\end{enumerate}

\subsection{Reliability Degree of Correlation Filter}
\label{Reliability Degree of Correlation Filter}

The peak value and the fluctuation of the response map can reveal the confidence degree about the tracking results to some extent. The ideal response map should have only one sharp peak and be smooth in all other areas when the detected target is extremely matched to the correct target. The sharper the correlation peaks are, the better the location accuracy is. Otherwise, the whole response map will fluctuate intensely, and its pattern is significantly different from normal response maps. If we continue to use uncertain samples to update the tracking model, it would be corrupted mostly.

Inspired by \cite{Wang2017Large}, we evaluate the stability of the $k$-th expert with two criteria. The first criterion is the maximum response score $F_{max}$ of the response map defined as
\begin{equation}\label{Eq:8}
  F_{max} = \max_{x',y'} \mathbf{\emph{S}}^{k}(x',y')
\end{equation}
where $\mathbf{\emph{S}}^{k}(x',y')$ is referred to Eq. \ref{Eq:4} and Eq. \ref{Eq:5}.

The second criterion is called average peak-to-correlation
energy (APCE) measure which is defined in Eq. \ref{Eq:6}.

\subsection{Multi-task Gaussian Particle Filter}
\label{Multi-task Gaussian Particle Filter}

The proposed multi-task Gaussian particle filter at time $t$ approximates the posterior mean $\bm{\mu}_{t}$ and covariance $\bm{\Sigma}_{t}$ of the unknown state variable $\mathbf{x}_{t}$ using Bayesian importance sampling.

We draw samples from the importance function $\pi(\cdot)$ at time $t$ using
\begin{equation}\label{Eq:9}
\pi(\mathbf{x}_{t}|\mathbf{y}_{0:t}) = \mathcal{N}(\mathbf{x}_{t};\bm{\mu}_{t},\mathbf{\Sigma}_{t}),
\end{equation}
and denote them as $\{\mathbf{x}_{t}^{j}\}^{M}_{j=1}$. Here, $\mathbf{y}_{0:t}$ is the observations over time $t$, and $\mathcal{N}(\cdot)$ represents a Gaussian function.
Note that, in Eq. \ref{Eq:9}, $\bm{\mu}_{t}$ is equal to Eq. \ref{Eq:7} and $\mathbf{\Sigma}_{1}$ is chosen based on prior information.

The respective weights are computed by
\begin{equation}\label{Eq:10}
  w_{t}^{j}=
  \frac{p(\mathbf{y}_{t}|\mathbf{x}_{t}^{j})
  \mathcal{N}(\mathbf{x}_{t}=\mathbf{x}_{t}^{j};\bm{\mu}_{t},\mathbf{\Sigma}_{t})}
  {\pi(\mathbf{x}_{t}^{j}|\mathbf{y}_{0:t})},
\end{equation}
where the distribution $p(\mathbf{y}_{t}|\mathbf{x}_{t}^{j})$ represents the observation equation $\mathbf{y}_{t}$ conditioned on the unknown state variable $\mathbf{x}_{t}^{j}$ at time $t$.

Eq. \ref{Eq:10} can be rewritten as follows from Eq. \ref{Eq:9}:
\begin{equation}\label{Eq:11}
w_{t}^{j} \propto  p(\mathbf{y}_{t}|\mathbf{x}_{t}^{j}).
\end{equation}

 Then, we set $p(\mathbf{y}_{t}|\mathbf{x}_{t}^{j}) = |f_{t}^{*}-f(\mathbf{x}_{t}^{j})|$, where $\left|\cdot \right|$ denotes the absolute value and $f_{t}$ is the function of features, where $f_{t}^{*}$ represents the features of the template at time $t$. Hence, each Gaussian particle weight can be calculated with
\begin{equation}\label{Eq:12}
w_{t}^{j}\propto |f_{t}^{*}-f(\mathbf{x}_{t}^{j})|
\end{equation}

Normalize the weights as
\begin{equation}\label{Eq:13}
\tilde{w}_{t}^{j} = \frac{w_{t}^{j}}{\sum_{j=1}^{M} w_{t}^{j}}.
\end{equation}

In this paper, we adopt HOG for tackling deformation variations and gray normalization of raw pixel values from sample images for handling illumination changes, and the $j$-th features are represented by $f^{j}_{t(hog)}$ and $f^{j}_{t(norm)}$ at time $t$, respectively . Hence, we get the corresponding weights $\tilde{w}_{t(hog)}^{j}$ and $\tilde{w}_{t(norm)}^{j}$.

For the multi-task Gaussian particle filter, we jointly define the similarities of respective samples as
\begin{equation}\label{Eq:14}
\bar{w}_{t}^{j} = \theta \cdot\tilde{w}_{t(hog)}^{j} + (1-\theta)\cdot\tilde{w}_{t(norm)}^{j},
\end{equation}
where learning rate parameter $\theta$ is a constant value in this paper.

The mean and covariance are estimated by
\begin{eqnarray}\label{Eq:15}
\bm{\mu}_{t} &=&  \sum_{j=1}^{M} \bar{w}_{t}^{j}\mathbf{x}_{t}^{j}, \\
\bm{\Sigma}_{t} &=& \sum_{j=1}^{M} \bar{w}_{t}^{j} (\bm{\mu}_{t}-\mathbf{x}_{t}^{j}) (\bm{\mu}_{t}-\mathbf{x}_{t}^{j})^{H},
\end{eqnarray}
where $H$ represents the Hermitian Matrix.

The maximun historical-moving-distance of the tracked target at time $t$ is defined as
\begin{equation}\label{Eq:17}
d_{t} = \max_{i=1}^{t} ( \left| \bm{\mu}_{i} - \bm{\mu}_{i-1}  \right|).
\end{equation}
Here, we set $d_{1} = 0$, and $ \left| \cdot  \right|$ denotes the function for absolution values;

\subsection{KCF-GPF Tracker}

Figure \ref{fig:flowchart} illustrates the flowchart of the proposed algorithm. Based on the structural correlation filter and a Gaussian particle filter, we propose a KCF-GPF tracker. The first step generates $K$ weak correlation filters. The second step is to make weak decisions via the weak structural correlation filter.s The third step is to evaluate the reliability degrees of weak decisions. Finally, the optimal decision is made using a multi-task Gaussian particle filter.

To update the KCF-GPF for visual tracking, we adopt an incremental strategy in the current frame to update the template in Eq. \ref{Eq:12} by
\begin{equation}\label{Eq:18}
  f_{t}^{*} = \rho \cdot f_{t-1}^{*} + (1-\rho)\cdot f_{t-1}(\bm{\mu}_{t-1}),
\end{equation}
where learning rate parameter $\rho$ is a constant value in this paper.

An overview of the proposed method is summarized in Algorithm \ref{Algorithm:1}.

\begin{algorithm}[htbp]
\label{Algorithm:1}
\caption{KCF-GPF tracking algorithm}
\KwIn{Frames {$\{\mathbf{I}_{t}\}_{1}^{T}$};}
\KwOut{Target location of each frame $\bm{\mu}_{t}$.}
\For{Time $t = 1:T $ }
{

    \eIf{$t=1$}
    {
        Initialize the target location $(x^{*}_{1},y^{*}_{1})$\;
        Crop interested image region\;
        Initiate $K$ weak correlation filters using Eq. \ref{Eq:3};
   }
    {
        Compute each correlation filter's response using Eq. \ref{Eq:4}\;
        Find target position predicted by each weak tracker using Eq. \ref{Eq:5}\;
        Evaluate the reliability degrees $APCE$ and $F_{max}$ via Eq. \ref{Eq:6} and Eq. \ref{Eq:8}\;
        \eIf{$\max\limits_{k=1}^{K} APCE$ $\mathrm {and}$ $\max\limits_{k=1}^{K} F_{max}$ $\mathrm {satisfy \ the \ condition}$}
        {Draw samples from Eq. \ref{Eq:9};}
        {Draw samples from Eq. \ref{Eq:9} where $\mathbf{\Sigma}_{t}=3$;}
        Compute the ultimate position using Eq. \ref{Eq:15}\;
        Update the template via Eq. \ref{Eq:18};
    }
}
\end{algorithm}

\section{Experiments}

\subsection{Experimental Setups}

\noindent \textbf{Implementation Details.} The conventional features used for KCF-GPF are composed
of HOG features and gray normalization features. Our tracker is implemented on MATLAB in a PC with a 2.80 GHz CPU and runs faster than 21 FPS in Table \ref{fig:OPE}. Our tracker requires few parameter settings, reported in Table \ref{fig:parameters setting}, where \lq padding' (referring to KCF \cite{Henriques2015High}) means the magnification of the image region samples relative to the target bounding box.
\begin{table}
  \centering
  \caption{Parameters of KCF-GPF}
  \begin{tabular}{|c|c|c|}
  \hline
Part of Track &parameters & values \\
\hline
\multirowcell{3}{KCF}&padding& 1.2\\
\cline{2-3}
                     &Feature bandwidth $\sigma$  & 0.5\\
\cline{2-3}
                     &Adaptation rate  & 0.045\\
\hline
\multirowcell{3}{GPF} & Sample numbers $N$ &  200\\
\cline{2-3}
                      & Learning rate $\theta$ (Eq. \ref{Eq:14})  & 0.65\\
\cline{2-3}
                      &  Learning rate $\rho$ (Eq. \ref{Eq:18}) &  0.8\\
\hline
\end{tabular}
\label{fig:parameters setting}
\end{table}
\\\\
\noindent \textbf{Datasets.} Our method is evaluated in the OTB-2013 dataset \cite{Wu2013Online} consisting of 50 sequences. The images are annotated with ground truth bounding boxes and 11 various visual attributes include scale variation, out of view, out-of-plane rotation, low resolution, in-plane rotation, illumination, motion blur, background clutter, occlusion, deformation, and fast motion.
\\\\
\noindent \textbf{Evaluation Metrics.} We compare the proposed method with the 16 state-of-the-art tracking methods using evaluation metrics and code provided by the respective benchmark dataset. For testing on OTB-2013, we employ the one-pass evaluation (OPE) and use two metrics: precision and success plots. The precision metric computes the rate of frames whose center location is within some certain distance from the ground truth location. The success metric computes the overlap ratio between the tracked and ground truth bounding boxes. In the legend, we report the area under curve (AUC) of success plot and precision score at a 20 pixel threshold (PS) corresponding to the one-pass evaluation for each tracking method.

\subsection{Comparison with State-of-the-Art}

\begin{table*}[htbp]
\scriptsize
\caption{Tracking results of all 17 evaluated trackers over all 50 sequences using OPE evaluation in the OTB-2013. The entries in {\color{red} red} denote the best results and the ones in {\color{blue} blue} indicate the second best.}
\begin{center}
\begin{tabular}{|c|c|c|c|c|c|c|c|c|c|c|}
\hline
\diagbox{OPE}{Trackers} & LMCF\cite{Wang2017Large} & CFNet\cite{Valmadre2017End} & CFN\cite{Choi2017Attentional} & CFN\underline{ } \cite{Choi2017Attentional} & CNT\cite{Zhang2016Robust} & BIT\cite{Cai2016BIT} & SINT\cite{Tao2016Siamese} & SCT\cite{Choi2016Visual} & Staple\cite{Bertinetto2015Staple} \\
\hline
precision & 0.842 & 0.803 & 0.813 & 0.784 & 0.723 & 0.816 & {\color{blue}0.851} & 0.836 & 0.793\\
\cline{1-10}
success & {\color{blue}0.800} & 0.775 & 0.675 & 0.630 & 0.656 & 0.745 &0.791 & 0.730& 0.754\\
\hline

\diagbox{OPE}{Trackers} & SiamFC\cite{Bertinetto2016Fully} & SRDCF\cite{Danelljan2015Learning} & DSST\cite{Danelljan2014Accurate} & MEEM\cite{Zhang2014MEEM} & KCF\cite{Henriques2015High} & TLD\cite{Kalal2012Tracking} & Struck\cite{Hare2012Struck} & KCF-GPF(ours)& mean FPS(ours) \\
\hline
precision & 0.809 & 0.838 & 0.740 & 0.840 & 0.740 & 0.608 & 0.656 &{\color{red}0.857}& \multirowcell{2}{21.3}\\
\cline{1-9}
success & 0.783 & 0.781 &0.670 & 0.706 & 0.623 & 0.521 & 0.559 & {\color{red}0.805}& \\
\hline
\end{tabular}
\end{center}
\label{fig:OPE}
\end{table*}

We evaluate KCF-GPF in the OTB-2013 dataset \cite{Wu2013Online} and compare it with 16 state-of-the-art trackers including LMCF \cite{Wang2017Large}, CFNet \cite{Valmadre2017End}, CFN \cite{Choi2017Attentional}, CFN\underline{ } \cite{Choi2017Attentional}, CNT \cite{Zhang2016Robust}, BIT \cite{Cai2016BIT}, SINT \cite{Tao2016Siamese}, SCT \cite{Choi2016Visual}, Staple \cite{Bertinetto2015Staple}, SiamFC \cite{Bertinetto2016Fully}, SRDCF \cite{Danelljan2015Learning}, DSST \cite{Danelljan2014Accurate}, MEEM \cite{Zhang2014MEEM}, KCF \cite{Henriques2015High}, TLD \cite{Kalal2012Tracking} and Struck \cite{Hare2012Struck}. Among them, LMCF, CFN, CFN\underline{ }, Staple, SRDCF, KCF, DSST, CFNet  and SCT are CF based algorithms; SiamFC, SINT, CFNet, CNT and BIT are convolutional networks based algorithms; MEEM is developed based on regression and multiple trackers; TLD is based on an ensemble classifier; and Struck is structured SVM based methods.

\begin{figure*}
\begin{center}
\includegraphics[width=0.24\linewidth]
                   {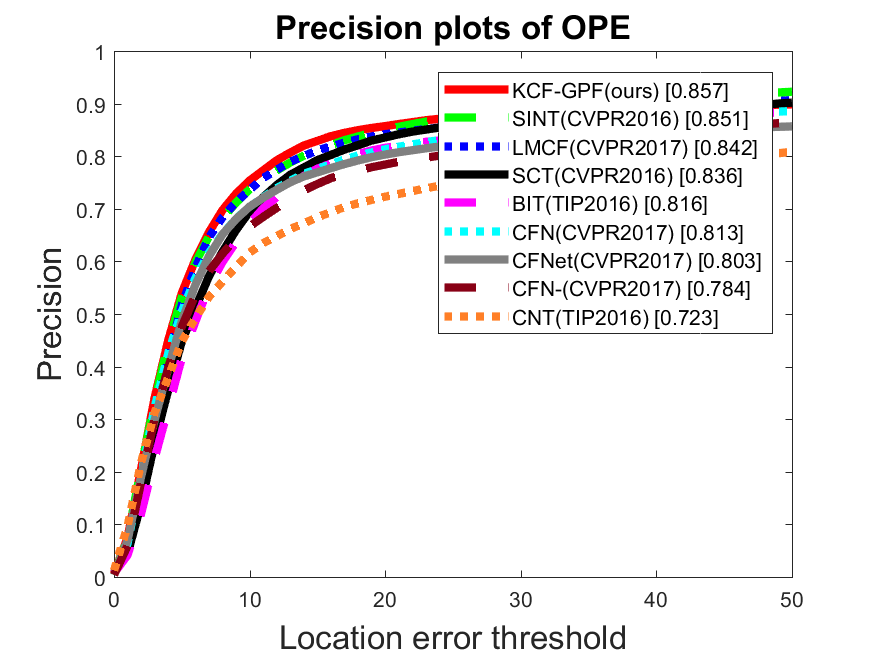}
\includegraphics[width=0.24\linewidth]
                   {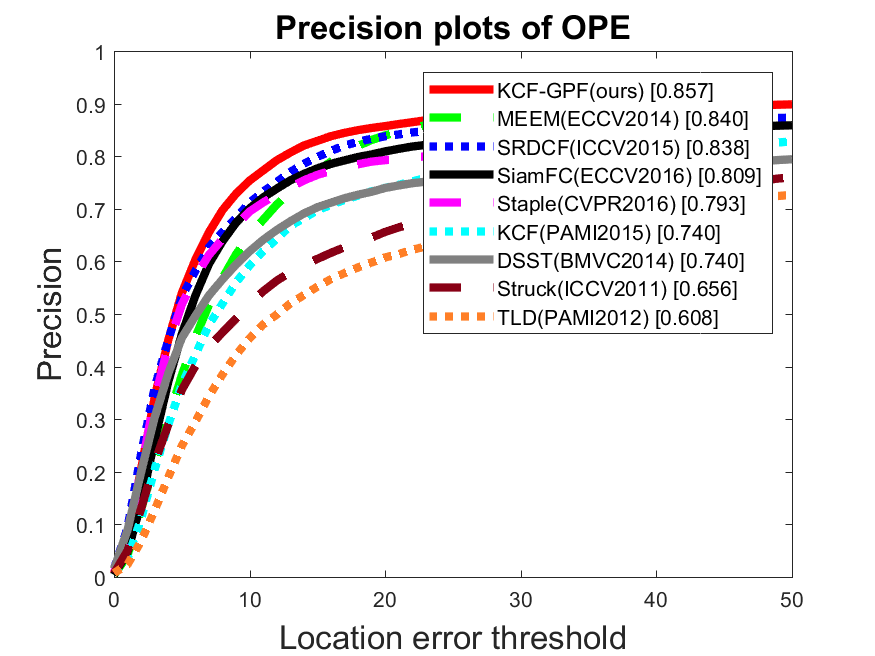}
\includegraphics[width=0.24\linewidth]
                   {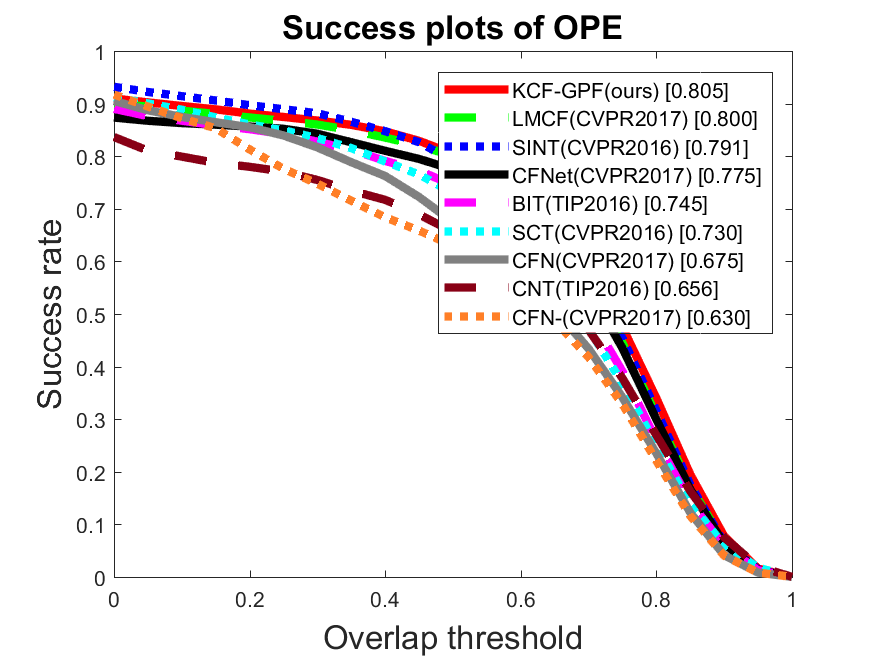}
\includegraphics[width=0.24\linewidth]
                   {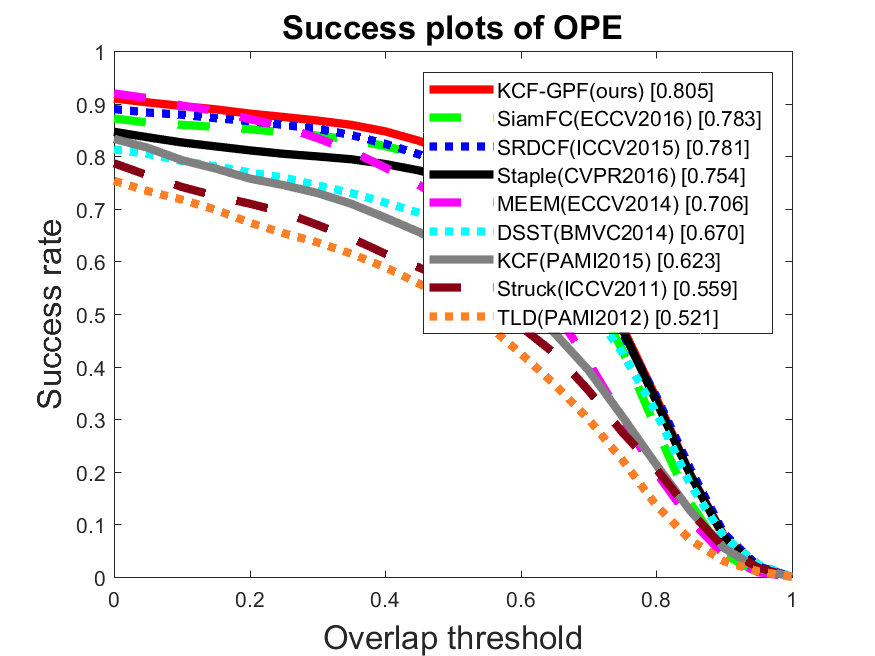}

\end{center}
\caption{Precision and success plots over all 50 sequences using OPE evaluation in the OTB-2013 dataset. The numbers in the legend indicate the average precision scores for precision plots and the average AUC scores for success plots. Our KCF-GPF method performs favorably against the state-of-the-art trackers.}
\label{fig:precision and success}
\end{figure*}

The characteristics and tracking results are summarized in Table \ref{fig:OPE}. The mean FPS here is estimated on all sequences in the OTB-2013 and achieves 21.3 fps satisfying the requirement of real-time capability. LMCF achieves the second best performance in terms of the success metric and SINT shows the second best performance in terms of precision metric. Figure \ref{fig:precision and success} illustrates the precision and success plots of all trackers under all challenging attributes in the OTB-2013. KCF-GPF is also superior to other up-to-date trackers with precision and success evaluation metrics in the OTB-2013 benchmark.

\begin{figure*}[htbp]
\begin{center}
\includegraphics[width=0.24\linewidth]
                   {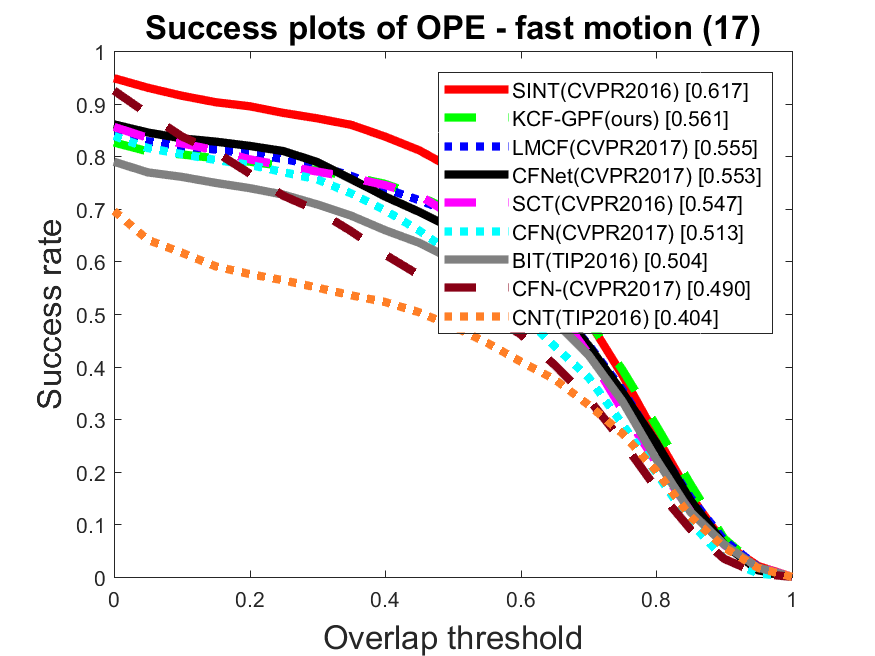}
\includegraphics[width=0.24\linewidth]
                   {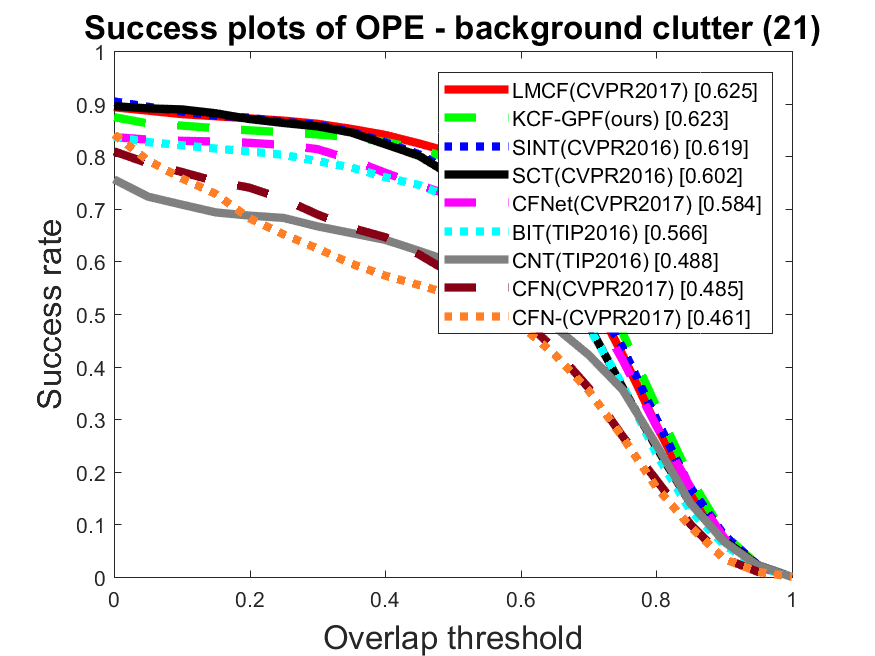}
\includegraphics[width=0.24\linewidth]
                   {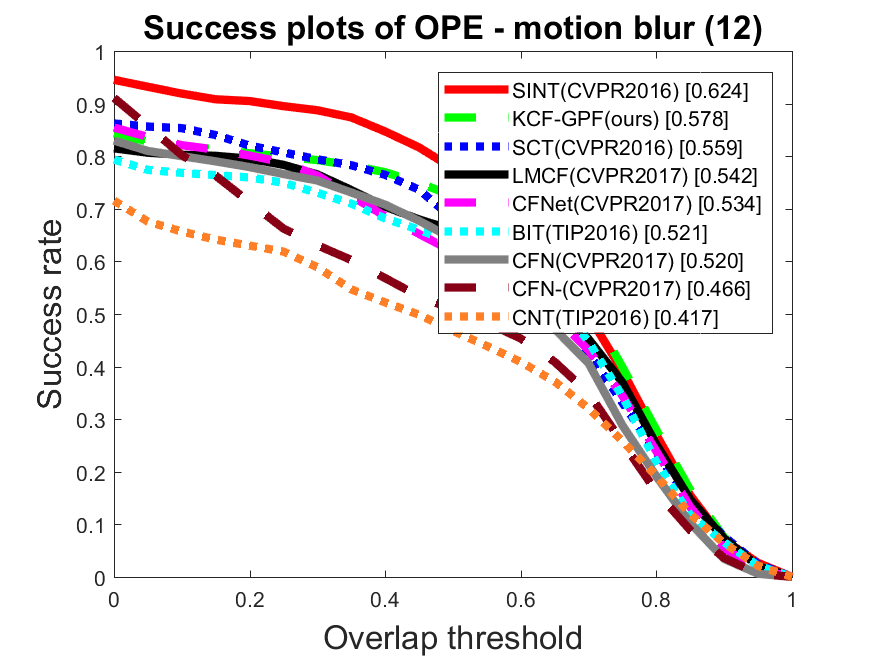}
\includegraphics[width=0.24\linewidth]
                   {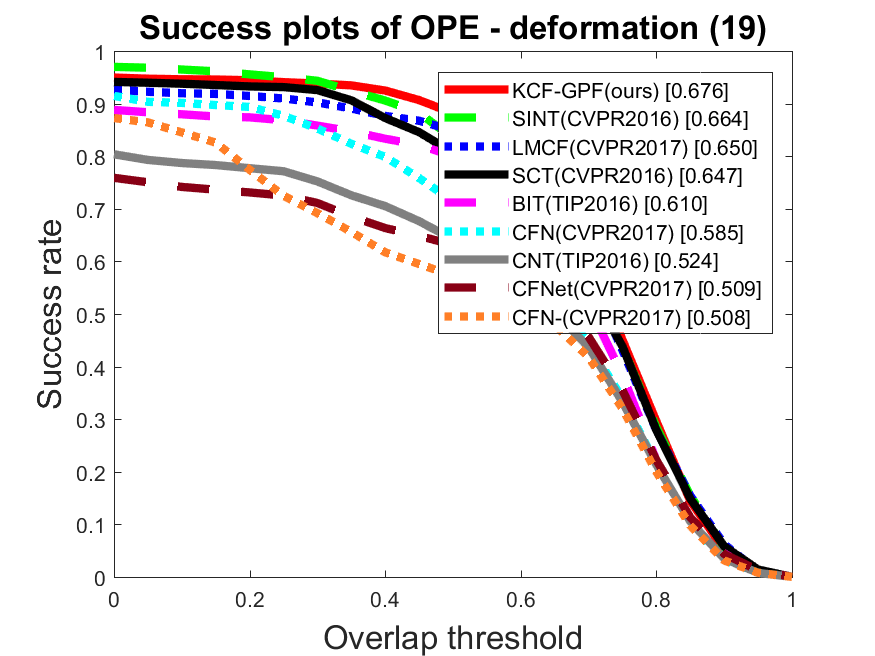}
\includegraphics[width=0.24\linewidth]
                   {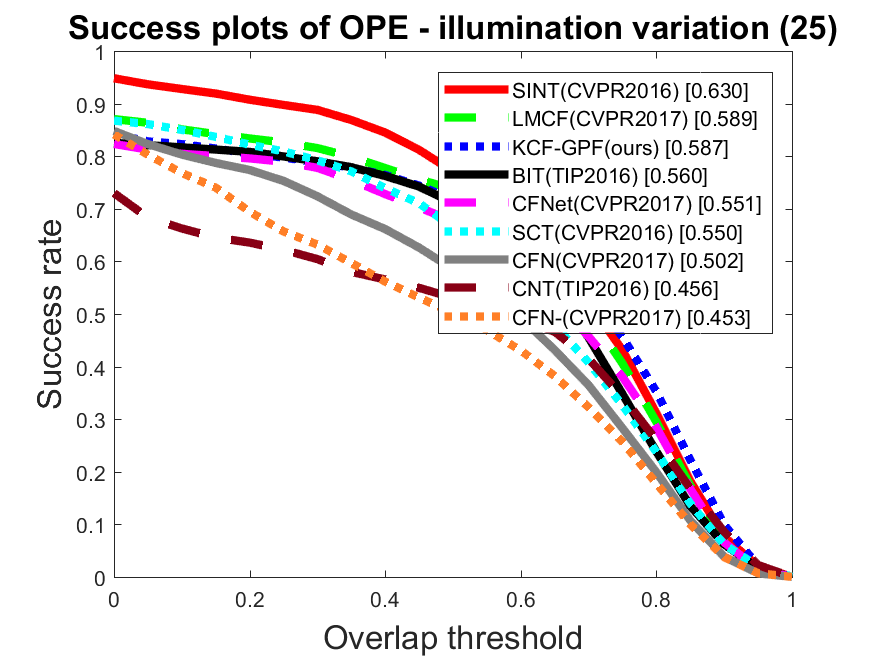}
\includegraphics[width=0.24\linewidth]
                   {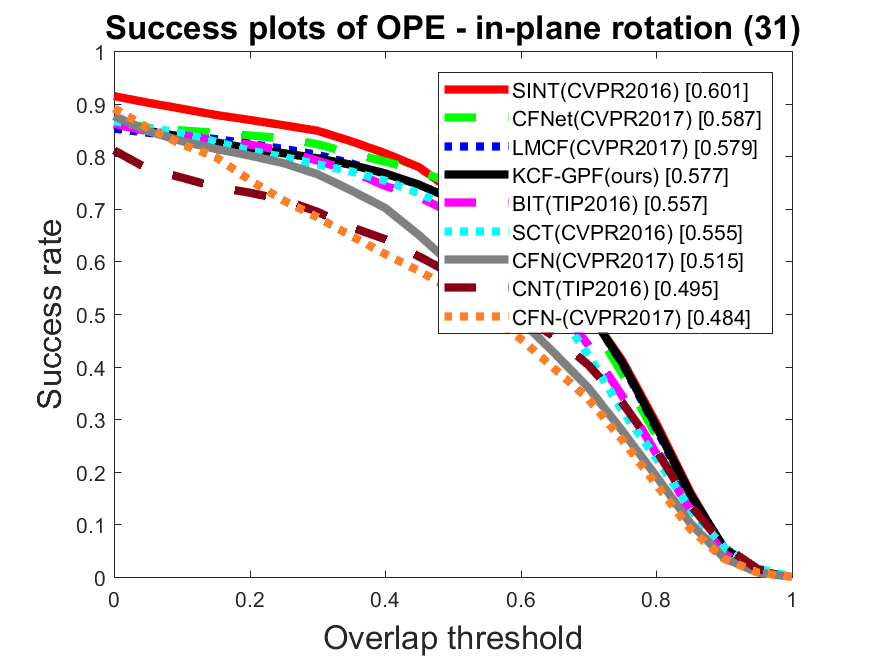}
\includegraphics[width=0.24\linewidth]
                   {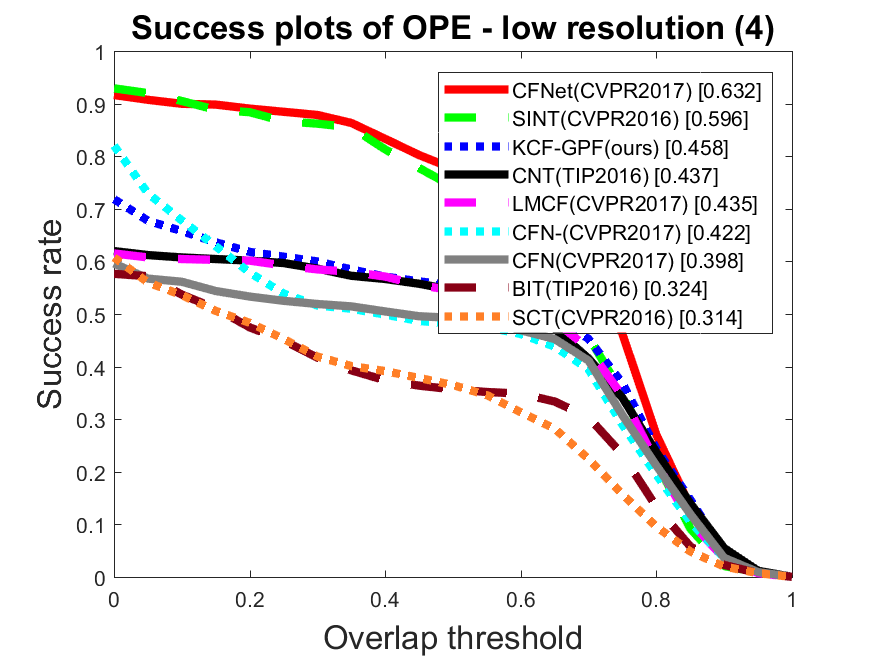}
\includegraphics[width=0.24\linewidth]
                   {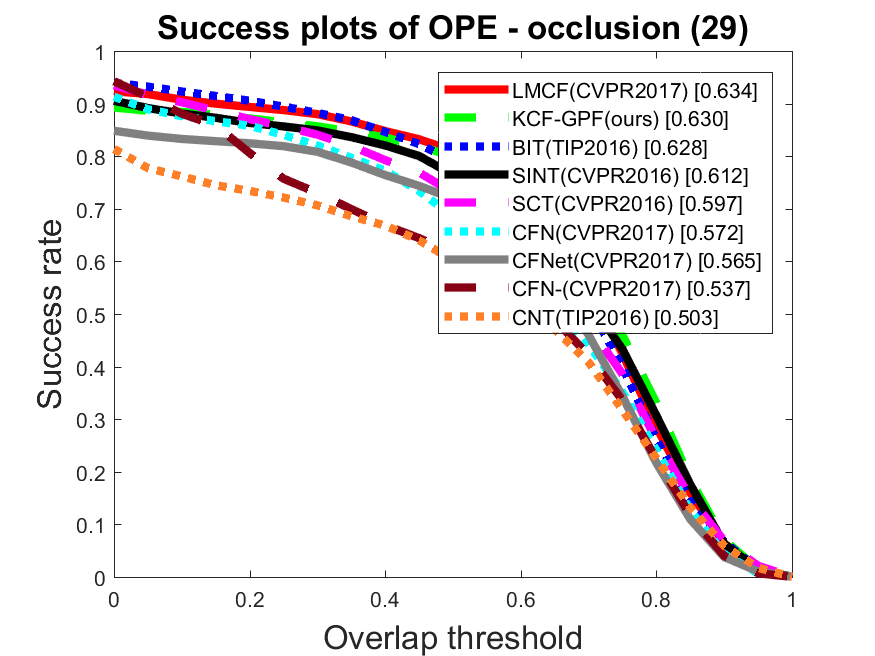}
\includegraphics[width=0.24\linewidth]
                   {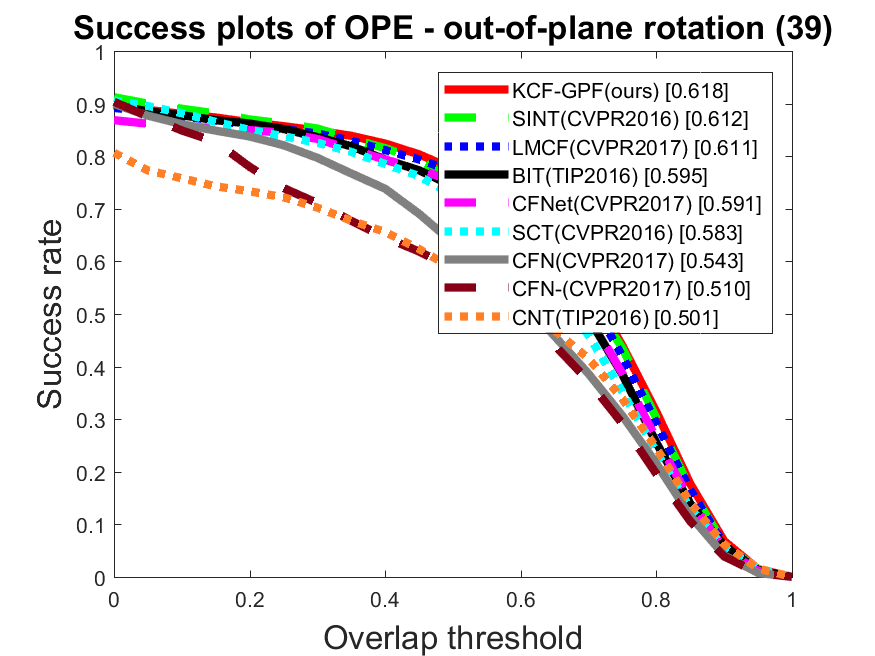}
\includegraphics[width=0.24\linewidth]
                   {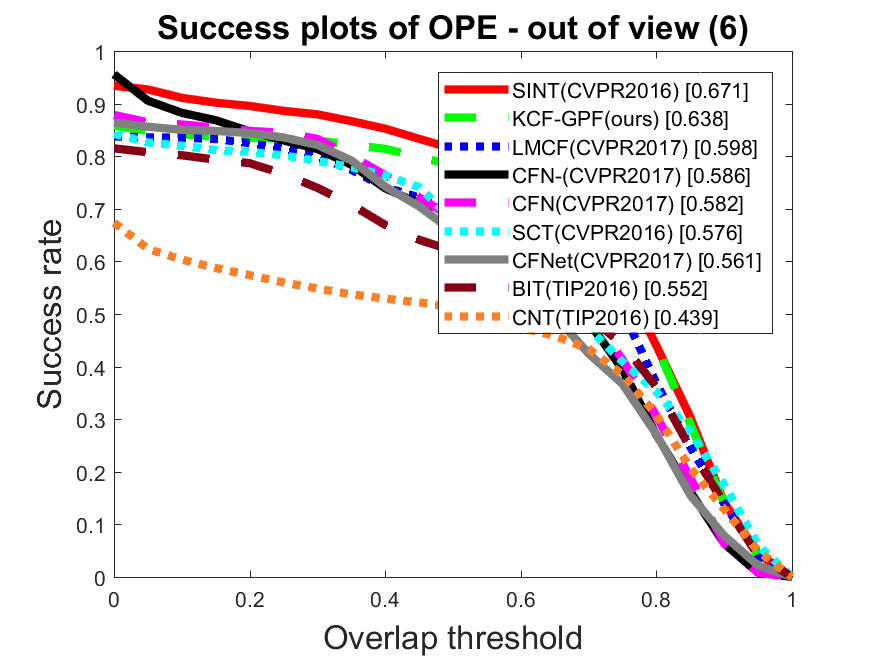}
\includegraphics[width=0.24\linewidth]
                   {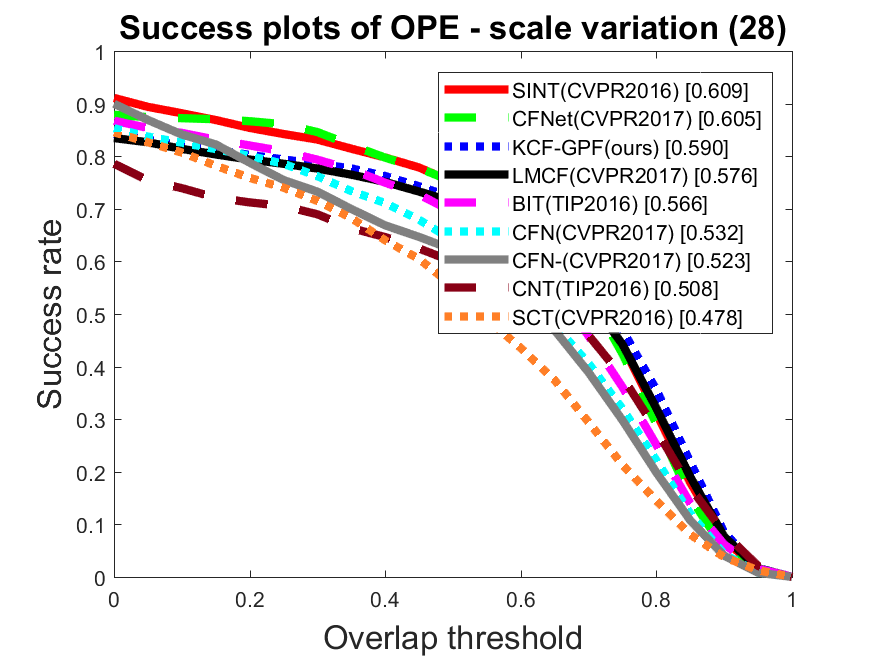}
\includegraphics[width=0.24\linewidth]
                   {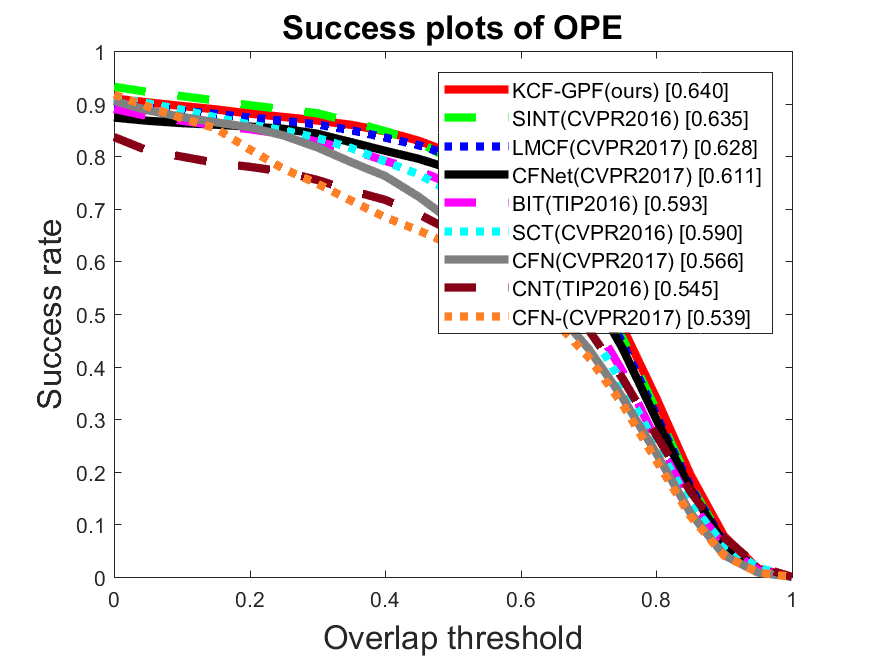}
\end{center}
\caption{Success plots over all 50 sequences using OPE evaluation in the OTB-2013 dataset. The evaluated trackers are LMCF, CFNet, CFN, CFN\underline{ }, CNT, BIT, SINT, SCT and KCF-GPF. All 11 tracking challenges include scale variation, out of view, out-of-plane rotation, low resolution, in-plane rotation, illumination, motion blur, background clutter, occlusion, deformation, and fast motion. The numbers in the legend indicate the average AUC scores for success plots. Our KCF-GPF method performs favorably against the state-of-the-art trackers.}
\label{fig:AUC2017-2016}
\end{figure*}

\begin{figure*}
\begin{center}
\includegraphics[width=0.24\linewidth]
                   {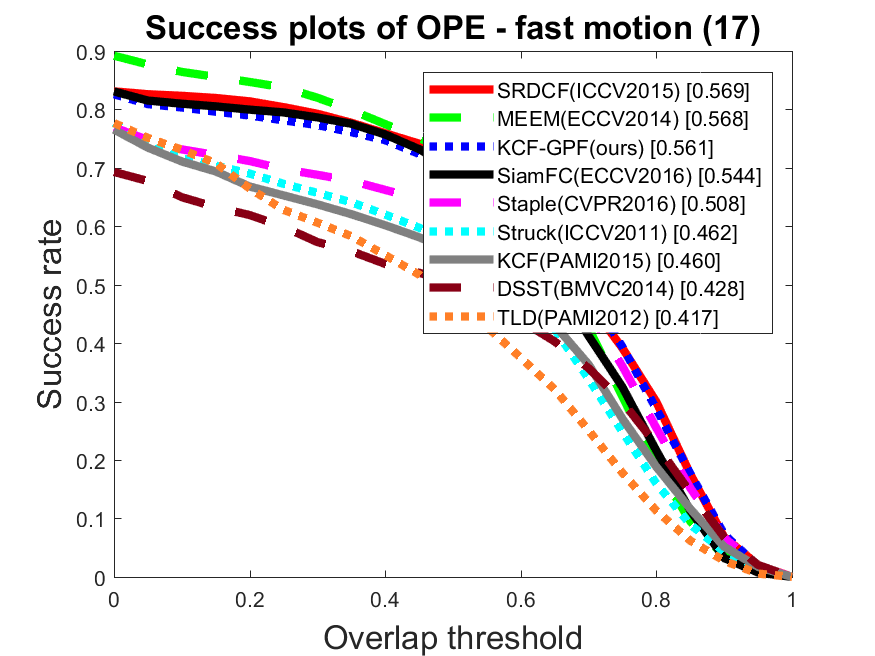}
\includegraphics[width=0.24\linewidth]
                   {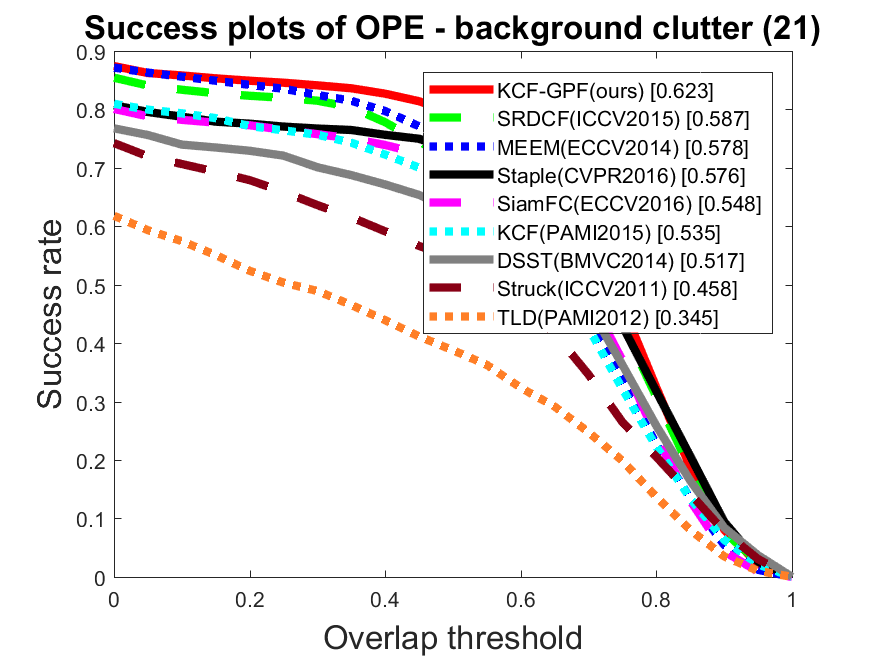}
\includegraphics[width=0.24\linewidth]
                   {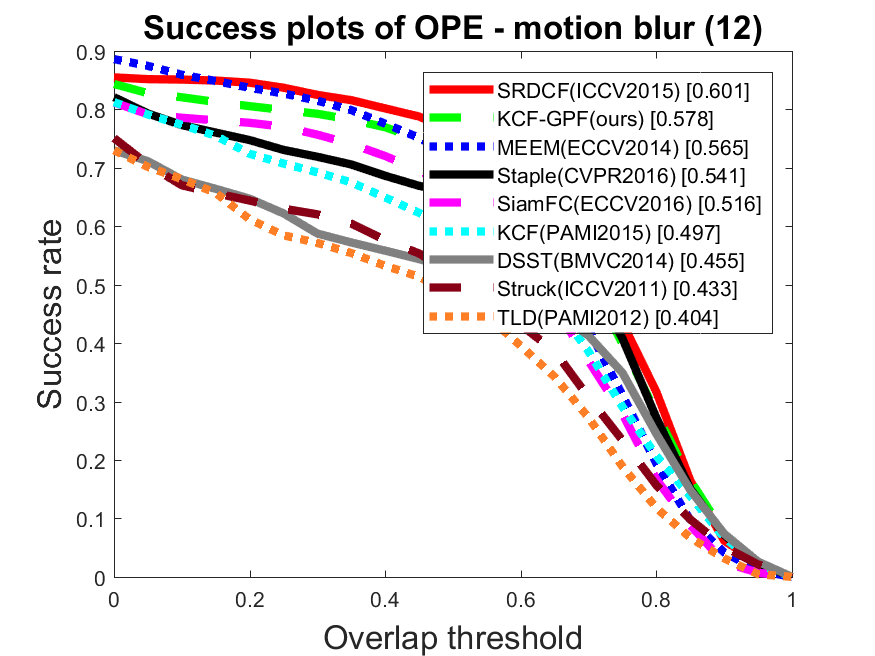}
\includegraphics[width=0.24\linewidth]
                   {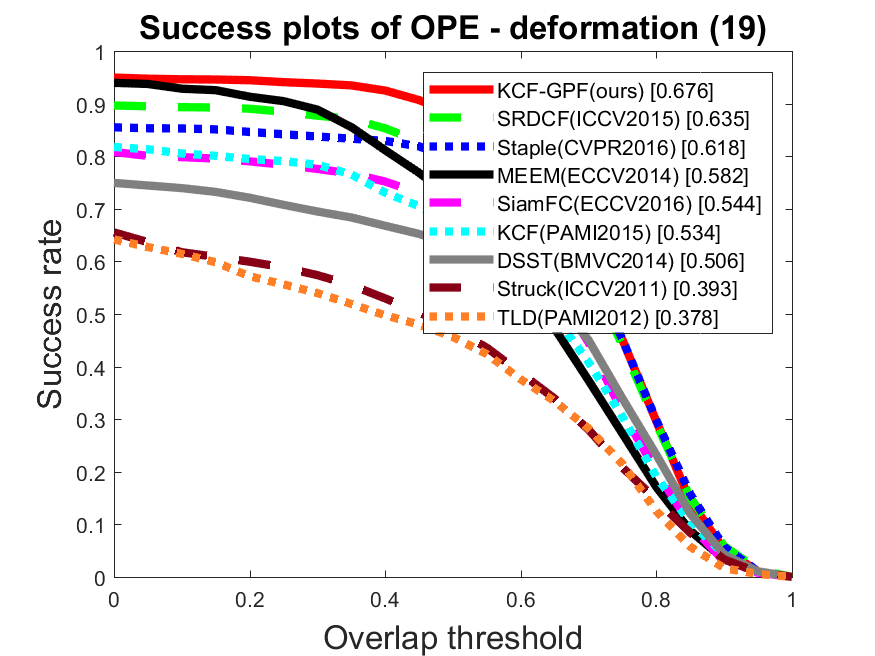}
\includegraphics[width=0.24\linewidth]
                   {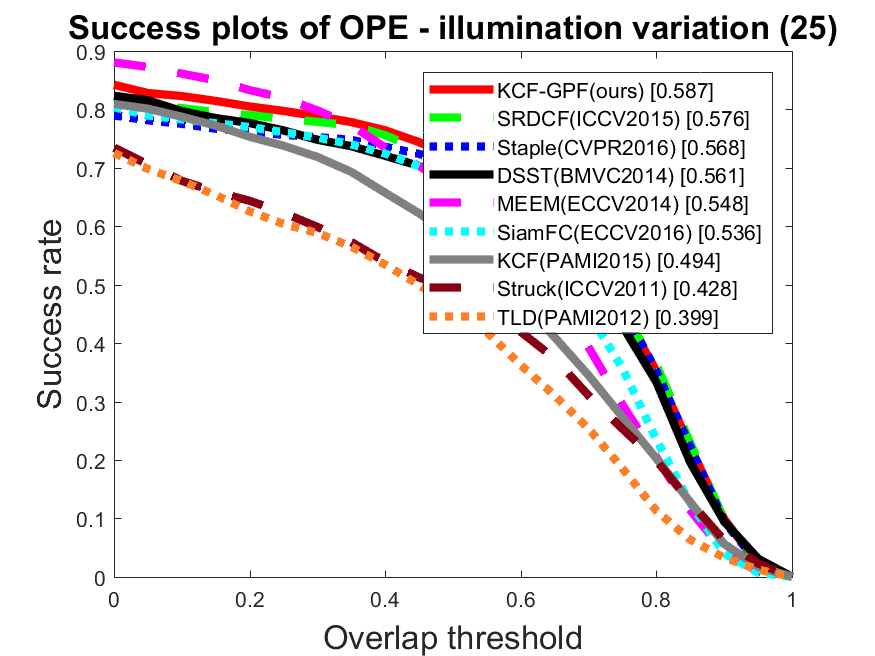}
\includegraphics[width=0.24\linewidth]
                   {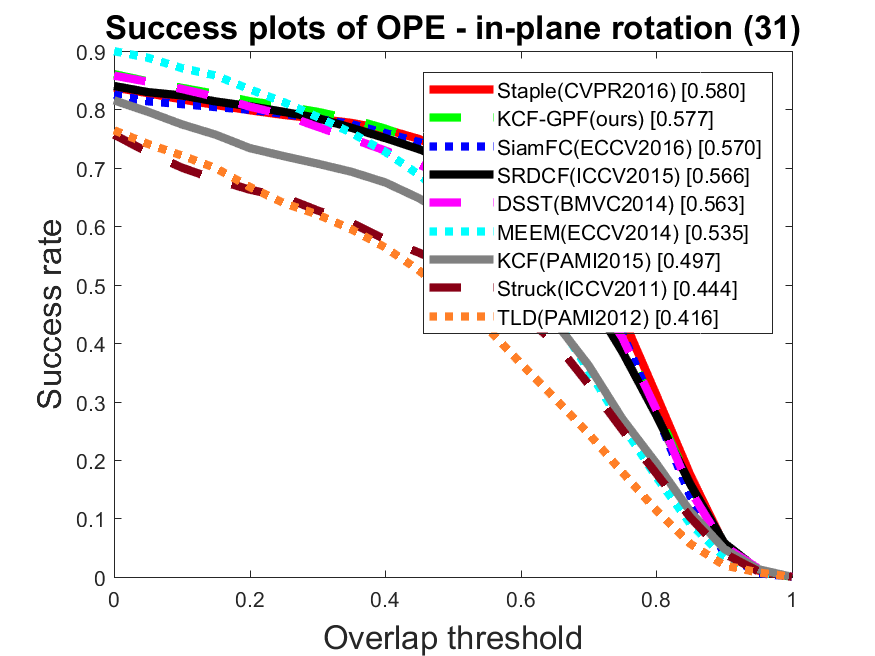}
\includegraphics[width=0.24\linewidth]
                   {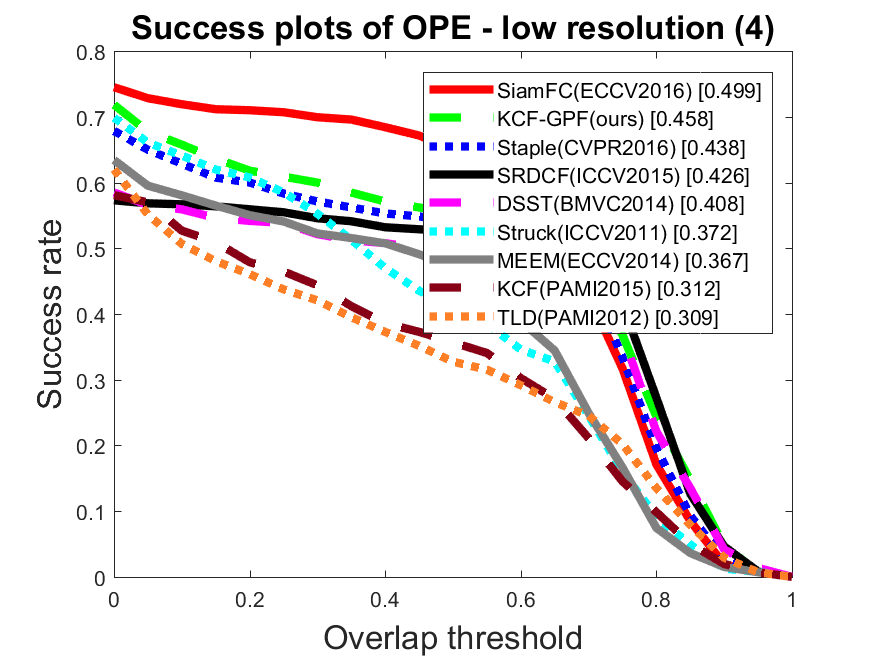}
\includegraphics[width=0.24\linewidth]
                   {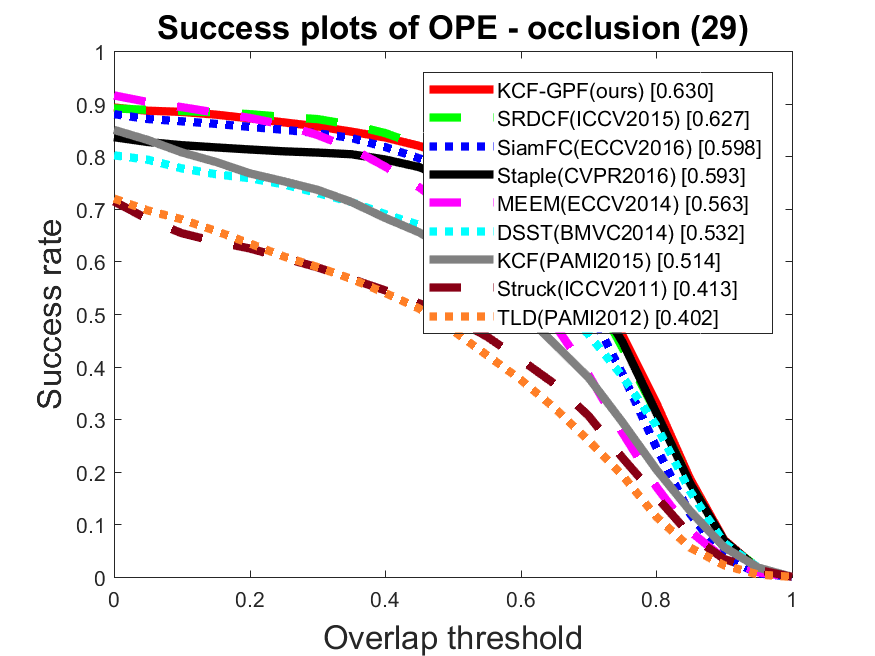}
\includegraphics[width=0.24\linewidth]
                   {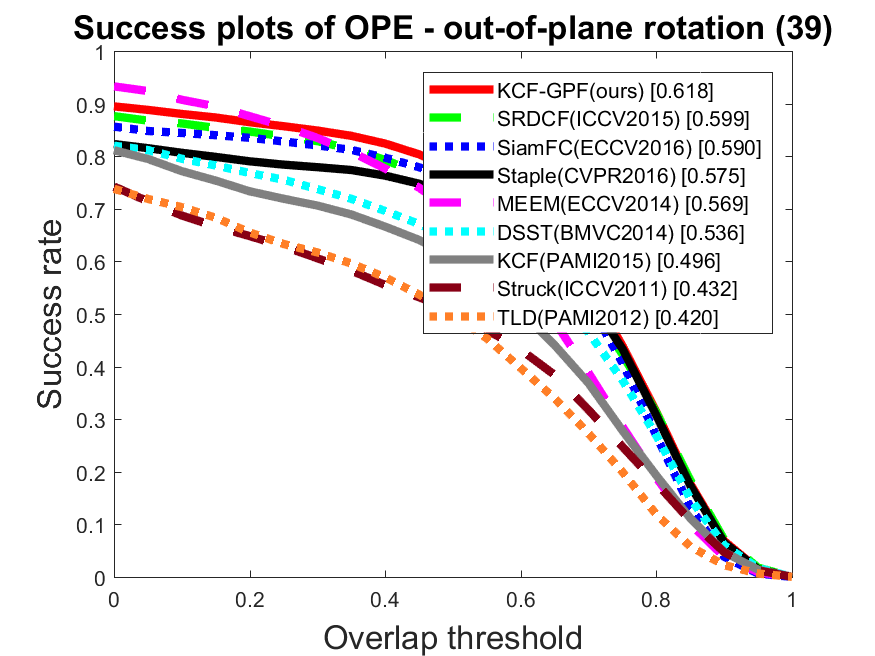}
\includegraphics[width=0.24\linewidth]
                   {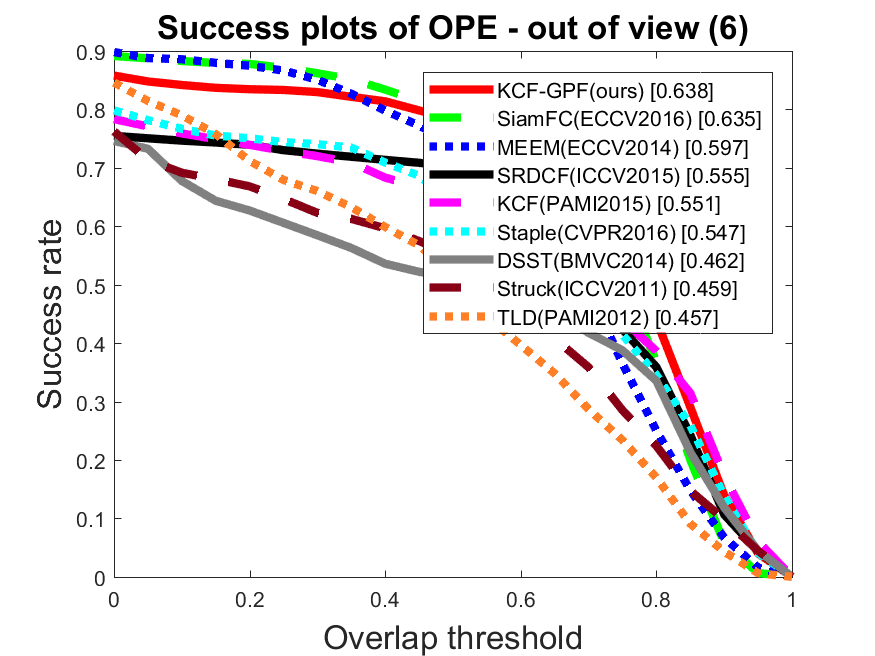}
\includegraphics[width=0.24\linewidth]
                   {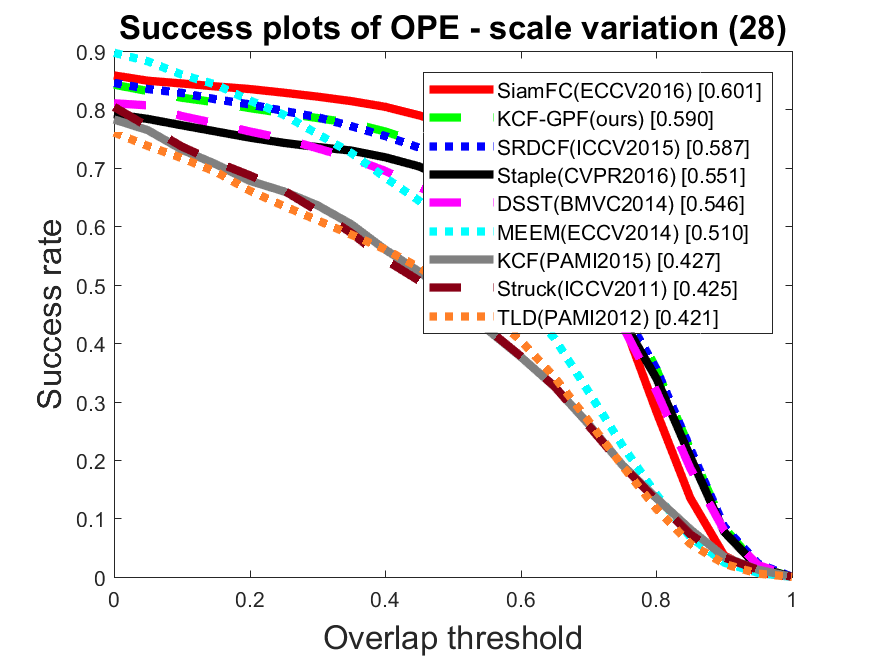}
\includegraphics[width=0.24\linewidth]
                   {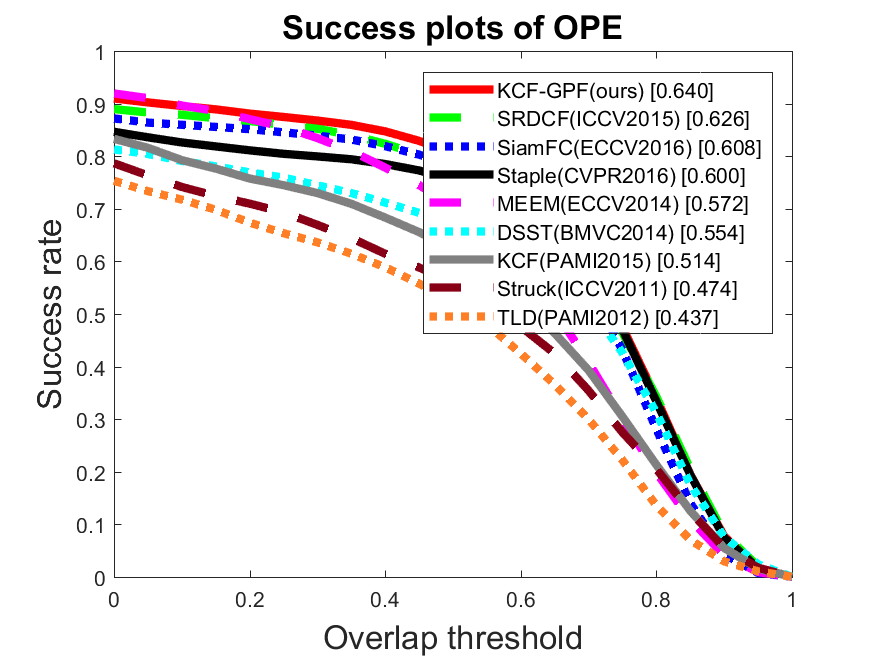}
\end{center}
\caption{Success plots over all 50 sequences using OPE evaluation in the OTB-2013 dataset. The evaluated trackers are Staple, SiamFC, SRDCF, DSST, MEEM, KCF, TLD, Struck and KCF-GPF. All 11 tracking challenges include scale variation, out of view, out-of-plane rotation, low resolution, in-plane rotation, illumination, motion blur, background clutter, occlusion, deformation, and fast motion. The numbers in the legend indicate the average AUC scores for success plots. Our KCF-GPF method performs favorably against the state-of-the-art trackers.}
\label{fig:AUC2016-2014}
\end{figure*}

For detailed analyses, we also evaluate KCF-GPF with these state-of-the-art trackers on various challenging attributes in the OTB-2013 benchmark dataset and the results are shown in Figure \ref{fig:AUC2017-2016} and Figure \ref{fig:AUC2016-2014}. The results demonstrate that KCF-GPF is ranked on top three in each attribute and achieves the best performances in the general success plots. Besides that, the proposed method outperforms other trackers in terms of deformation and out-of-plane rotation attributes.

\subsection{Qualitative Comparison}

\begin{figure*}
  \centering
  % Requires \usepackage{graphicx}
  \includegraphics[width=6.1in]{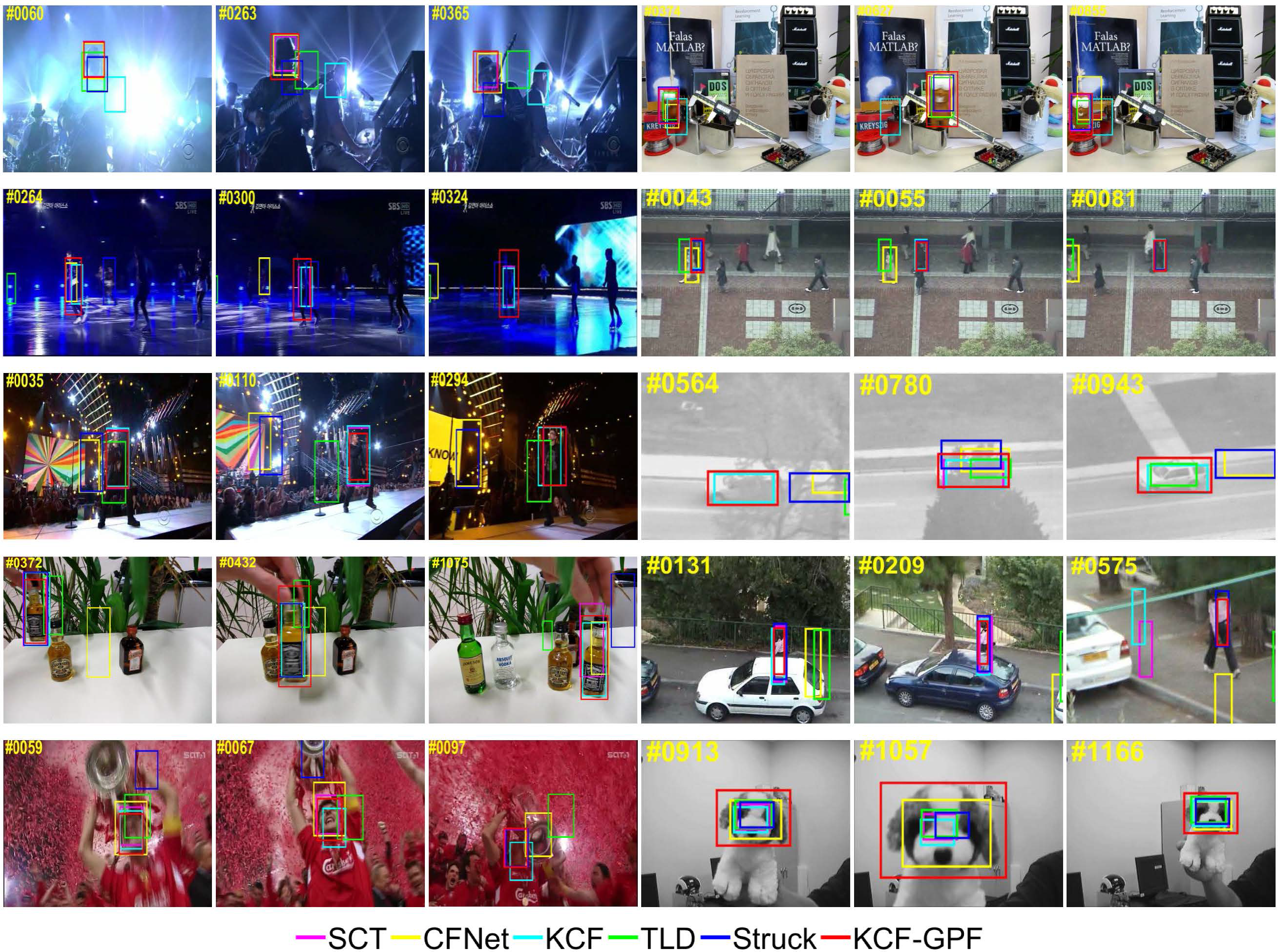}\\

  \caption{Comparisons of the proposed tracker with the state-of-the-art trackers (SCT \cite{Choi2016Visual}, CFNet \cite{Valmadre2017End}, KCF \cite{Henriques2015High}, \cite{Kalal2012Tracking} and Struck \cite{Hare2012Struck}) in our evaluation on 10 challenging sequences (from left to right and top to down are \textbf{Shaking}, \textbf{Lemming}, \textbf{Skating1}, \textbf{Subway}, \textbf{Singer2}, \textbf{Suv}, \textbf{Liquor}, \textbf{Woman}, \textbf{Soccer}, \textbf{Dog1}, respectively).}\label{fig:results}
\end{figure*}

To demonstrate the effect of the proposed KCF-GPF tracking algorithm, we make a qualitative comparison with above state-of-the-art trackers in the OTB-2013 benchmark dataset with 11 different attributes. As shown in Figure \ref{fig:results}, all trackers perform well overall, but the existing trackers have the following drawbacks. The SCT does not perform well under scale variations (Liquor, Woman, and Dog1). The CFNet cannot handle occlusion (Lemming, Skating1, Subway, Singer2, Suv, Liquor, Woman and Soccer), deformation(Skating1, Subway, Singer2, Suv and Woman) ,out-of-plane rotation (Lemming, Skating1, Singer2, Liquor, Woman and Soccer) and background clutters (Skating1, Subway, Singer2, Suv, Liquor and Soccer). The KCF drifts when there are illumination variations (Shaking, Lemming and Woman), scale variations (Shaking, Lemming, Woman and Dog1), out-of-plane rotations (Shaking, Lemming, Woman and Soccer), and fast motions (Woman and Soccer). The TLD and Struck methods drift when target objects undergo illumination changes (Shaking, Skating1, Singer2 and Soccer), heavy occlusion (Lemming, Subway, Singer2, Suv, Liquor, Woman and Soccer) and scale variations (Lemming and Dog1). Overall, the proposed KCF-GPF tracker performs the best against the existing trackers in tracking objects on these challenging sequences.

%------------------------------------------------------------------------
\section{Conclusion}

In this paper, we have proposed a novel tracker combining multiple structural correlation filters with a multi-task gaussian particle filter, namely KCF-GPF, to construct a strong tracker for ensemble tracking. The proposed method takes multiple correlation filters as weak expert trackers, and exploits spatial-geometric relations between target locations in consecutive frames to provide weak decisions in a reliable search scope. The reliability degrees of weak decisions are introduced in experiments for the GPF to make a strong decision. As a result, it not only has the advantages of the existing correlation filter trackers, such as, computational efficiency and robustness, but also can deal with scale variations by the sampling strategy of a GPF. The proposed KCF-GPF tracking algorithm outperforms 16 state-of-the-art methods over all 50 sequences in the OTB-2013 benchmark in terms of qualitative and quantitative evaluations.

{\small
\bibliographystyle{ieee}
\bibliography{egbib}

\begin{thebibliography}{10}\itemsep=-1pt

\bibitem{Avidan2007Ensemble}
S.~Avidan.
\newblock Ensemble tracking.
\newblock {\em IEEE Transactions on Pattern Analysis and Machine Intelligence},
  29(2):261--271, 2007.

\bibitem{Bai2013Randomized}
Q.~Bai, Z.~Wu, S.~Sclaroff, M.~Betke, and C.~Monnier.
\newblock Randomized ensemble tracking.
\newblock In {\em IEEE International Conference on Computer Vision}, pages
  2040--2047, 2013.

\bibitem{Bertinetto2015Staple}
L.~Bertinetto, J.~Valmadre, S.~Golodetz, O.~Miksik, and P.~H.~S. Torr.
\newblock Staple: Complementary learners for real-time tracking.
\newblock 38(2):1401--1409, 2015.

\bibitem{Bertinetto2016Fully}
L.~Bertinetto, J.~Valmadre, J.~F. Henriques, A.~Vedaldi, and P.~H.~S. Torr.
\newblock Fully-convolutional siamese networks for object tracking.
\newblock pages 850--865, 2016.

\bibitem{Carlo1997Dynamic}
C.~Berzuini, N.~G. Best, W.~R. Gilks, and C.~Larizza.
\newblock Dynamic conditional independence models and markov chain monte carlo
  methods.
\newblock {\em Journal of the American Statistical Association},
  92(440):1403--1412, 1997.

\bibitem{Bolme2010Visual}
D.~S. Bolme, J.~R. Beveridge, B.~A. Draper, and Y.~M. Lui.
\newblock Visual object tracking using adaptive correlation filters.
\newblock In {\em Computer Vision and Pattern Recognition}, pages 2544--2550,
  2010.

\bibitem{Boyd2011Distributed}
S.~Boyd, N.~Parikh, E.~Chu, B.~Peleato, and J.~Eckstein.
\newblock Distributed optimization and statistical learning via the alternating
  direction method of multipliers.
\newblock {\em Foundations and Trends in Machine Learning}, 3(1):1--122, 2011.

\bibitem{Cai2016BIT}
B.~Cai, X.~Xu, X.~Xing, K.~Jia, J.~Miao, and D.~Tao.
\newblock Bit: Biologically inspired tracker.
\newblock {\em IEEE Transactions on Image Processing}, 25(3):1327--1339, 2016.

\bibitem{Chaudhuri2009A}
K.~Chaudhuri, Y.~Freund, and D.~Hsu.
\newblock A parameter-free hedging algorithm.
\newblock {\em Computer Science}, pages 297--305, 2009.

\bibitem{Choi2016Visual}
J.~Choi, H.~J. Chang, J.~Jeong, Y.~Demiris, and Y.~C. Jin.
\newblock Visual tracking using attention-modulated disintegration and
  integration.
\newblock In {\em Computer Vision and Pattern Recognition}, pages 4321--4330,
  2016.

\bibitem{Choi2017Attentional}
J.~Choi, H.~J. Chang, S.~Yun, T.~Fischer, Y.~Demiris, and Y.~C. Jin.
\newblock Attentional correlation filter network for adaptive visual tracking.
\newblock In {\em IEEE Conference on Computer Vision and Pattern Recognition},
  2017.

\bibitem{Danelljan2014Accurate}
M.~Danelljan, G.~Hager, F.~S. Khan, and M.~Felsberg.
\newblock Accurate scale estimation for robust visual tracking.
\newblock In {\em British Machine Vision Conference}, pages 65.1--65.11, 2014.

\bibitem{Danelljan2015Learning}
M.~Danelljan, G.~Hager, F.~S. Khan, and M.~Felsberg.
\newblock Learning spatially regularized correlation filters for visual
  tracking.
\newblock In {\em IEEE International Conference on Computer Vision}, pages
  4310--4318, 2015.

\bibitem{Danelljan2016Discriminative}
M.~Danelljan, G.~Hager, F.~S. Khan, and M.~Felsberg.
\newblock Discriminative scale space tracking.
\newblock {\em IEEE Transactions on Pattern Analysis and Machine Intelligence},
  39(8):1561--1575, 2016.

\bibitem{Freund1995A}
Y.~Freund and R.~E. Schapire.
\newblock A desicion-theoretic generalization of on-line learning and an
  application to boosting.
\newblock In {\em European Conference on Computational Learning Theory}, pages
  23--37, 1995.

\bibitem{Gustafsson2002Particle}
F.~Gustafsson, F.~Gunnarsson, N.~Bergman, U.~Forssell, J.~Jansson, R.~Karlsson,
  and P.~J. Nordlund.
\newblock Particle filters for positioning, navigation, and tracking.
\newblock {\em IEEE Transactions on Signal Processing}, 50(2):425--437, 2002.

\bibitem{Hare2012Struck}
S.~Hare, A.~Saffari, and P.~H.~S. Torr.
\newblock Struck: Structured output tracking with kernels.
\newblock In {\em IEEE International Conference on Computer Vision}, pages
  263--270, 2012.

\bibitem{Harrison1976Bayesian}
P.~J. Harrison and C.~F. Stevens.
\newblock Bayesian forecasting. discussion.
\newblock {\em Journal of the Royal Statistical Society}, 38, 1976.

\bibitem{Henriques2015High}
J.~F. Henriques, C.~Rui, P.~Martins, and J.~Batista.
\newblock High-speed tracking with kernelized correlation filters.
\newblock {\em IEEE Transactions on Pattern Analysis and Machine Intelligence},
  37(3):583--596, 2015.

\bibitem{Jazwinski1970Stochastic}
Jazwinski and AndrewH.
\newblock {\em Stochastic processes and filtering theory}.
\newblock Academic Press,, 1970.

\bibitem{Julier2004Unscented}
S.~J. Julier and J.~K. Uhlmann.
\newblock Unscented filtering and nonlinear estimation.
\newblock {\em Proceedings of the IEEE}, 92(3):401--422, 2004.

\bibitem{Kalal2012Tracking}
Z.~Kalal, K.~Mikolajczyk, and J.~Matas.
\newblock Tracking-learning-detection.
\newblock {\em IEEE Transactions on Pattern Analysis and Machine Intelligence},
  34(7):1409, 2012.

\bibitem{Kotecha2003Gaussian}
J.~H. Kotecha and P.~M. Djuric.
\newblock Gaussian particle filtering.
\newblock {\em IEEE Transactions on Signal Processing}, 51(10):2592--2601,
  2003.

\bibitem{Li2015Reliable}
Y.~Li, J.~Zhu, and S.~C.~H. Hoi.
\newblock Reliable patch trackers: Robust visual tracking by exploiting
  reliable patches.
\newblock In {\em Computer Vision and Pattern Recognition}, pages 353--361,
  2015.

\bibitem{Liu1998Sequential}
J.~S. Liu and R.~Chen.
\newblock Sequential monte carlo methods for dynamic systems.
\newblock {\em Journal of the American Statistical Association},
  93(443):1032--1044, 1998.

\bibitem{Liu2001A}
J.~S. Liu, R.~Chen, and T.~Logvinenko.
\newblock {\em A Theoretical Framework for Sequential Importance Sampling with
  Resampling}.
\newblock Springer New York, 2001.

\bibitem{Liu2016Structural}
S.~Liu, T.~Zhang, X.~Cao, and C.~Xu.
\newblock Structural correlation filter for robust visual tracking.
\newblock In {\em Computer Vision and Pattern Recognition}, pages 4312--4320,
  2016.

\bibitem{Liu2015Real}
T.~Liu, G.~Wang, and Q.~Yang.
\newblock Real-time part-based visual tracking via adaptive correlation
  filters.
\newblock In {\em Computer Vision and Pattern Recognition}, pages 4902--4912,
  2015.

\bibitem{Ma2015Hierarchical}
C.~Ma, J.~B. Huang, X.~Yang, and M.~H. Yang.
\newblock Hierarchical convolutional features for visual tracking.
\newblock In {\em IEEE International Conference on Computer Vision}, pages
  3074--3082, 2015.

\bibitem{Ma2015Long}
C.~Ma, X.~Yang, C.~Zhang, and M.~H. Yang.
\newblock Long-term correlation tracking.
\newblock In {\em IEEE Conference on Computer Vision and Pattern Recognition},
  pages 5388--5396, 2015.

\bibitem{Mendel1980Optimal}
J.~Mendel.
\newblock Optimal filtering.
\newblock {\em IEEE Transactions on Automatic Control}, 25(3):615--616, 1980.

\bibitem{Merwe2001The}
R.~V.~D. Merwe, A.~Doucet, N.~D. Freitas, and E.~Wan.
\newblock The unscented particle filter.
\newblock {\em Advances in Neural Information Processing Systems}, 13:584--590,
  2001.

\bibitem{Nummiaro2003An}
K.~Nummiaro, E.~Koller-Meier, and L.~V. Gool.
\newblock An adaptive color-based particle filter.
\newblock {\em Image and Vision Computing}, 21(1):99--110, 2003.

\bibitem{Qi2016Hedged}
Y.~Qi, S.~Zhang, L.~Qin, H.~Yao, Q.~Huang, J.~Lim, and M.~H. Yang.
\newblock Hedged deep tracking.
\newblock In {\em Computer Vision and Pattern Recognition}, pages 4303--4311,
  2016.

\bibitem{Rui2012Exploiting}
C.~Rui, P.~Martins, and J.~Batista.
\newblock Exploiting the circulant structure of tracking-by-detection with
  kernels.
\newblock In {\em European Conference on Computer Vision}, pages 702--715,
  2012.

\bibitem{Smeulders2014Visual}
A.~W.~M. Smeulders, D.~M. Chu, R.~Cucchiara, S.~Calderara, A.~Dehghan, and
  M.~Shah.
\newblock Visual tracking: An experimental survey.
\newblock {\em IEEE Transactions on Pattern Analysis and Machine Intelligence},
  36(7):1442--1468, 2014.

\bibitem{Sorenson1988Recursive}
H.~W. Sorenson.
\newblock Recursive estimation for nonlinear dynamic systems.
\newblock {\em Bayesian Analysis of Time}, 1988.

\bibitem{Tao2016Siamese}
R.~Tao, E.~Gavves, and A.~W.~M. Smeulders.
\newblock Siamese instance search for tracking.
\newblock In {\em Computer Vision and Pattern Recognition}, pages 1420--1429,
  2016.

\bibitem{Valmadre2017End}
J.~Valmadre, L.~Bertinetto, J.~F. Henriques, A.~Vedaldi, and P.~H.~S. Torr.
\newblock End-to-end representation learning for correlation filter based
  tracking.
\newblock 2017.

\bibitem{Wang2017Large}
M.~Wang, Y.~Liu, and Z.~Huang.
\newblock Large margin object tracking with circulant feature maps.
\newblock 2017.

\bibitem{Wang2014Ensemble}
N.~Wang and D.~Y. Yeung.
\newblock Ensemble-based tracking: aggregating crowdsourced structured time
  series data.
\newblock In {\em International Conference on International Conference on
  Machine Learning}, pages II--1107, 2014.

\bibitem{Welch2010An}
G.~Welch and G.~Bishop.
\newblock An introduction to the kalman filter.
\newblock {\em University of North Carolina at Chapel Hill}, 8(7):127--132,
  2010.

\bibitem{Wu2013Online}
Y.~Wu, J.~Lim, and M.~H. Yang.
\newblock Online object tracking: A benchmark.
\newblock In {\em IEEE Conference on Computer Vision and Pattern Recognition},
  pages 2411--2418, 2013.

\bibitem{Yilmaz2006Object}
A.~Yilmaz.
\newblock Object tracking: A survey.
\newblock {\em Acm Computing Surveys}, 38(4):13, 2006.

\bibitem{Zhang2014MEEM}
J.~Zhang, S.~Ma, and S.~Sclaroff.
\newblock Meem: Robust tracking via multiple experts using entropy
  minimization.
\newblock In {\em European Conference on Computer Vision}, pages 188--203,
  2014.

\bibitem{Zhang2016Robust}
K.~Zhang, Q.~Liu, Y.~Wu, and M.~H. Yang.
\newblock Robust visual tracking via convolutional networks without training.
\newblock {\em IEEE Transactions on Image Processing A Publication of the IEEE
  Signal Processing Society}, 25(4):1779, 2016.

\end{thebibliography}
}

\end{document}